\documentclass{article}
\usepackage{spconf,amsmath,graphicx}

\usepackage[english]{babel}
\usepackage[T1]{fontenc}  
\usepackage[latin1]{inputenc} 
\usepackage{url}
\usepackage{graphicx}
\usepackage{subfigure}
\usepackage{xcolor}
\usepackage{amsmath,amsfonts,amssymb}
\usepackage{booktabs}
\usepackage{epsfig}
\usepackage{pstricks}
\usepackage{pst-node}
\usepackage{pst-grad}
\usepackage{ifthen}
\usepackage{algorithm}
\usepackage{algorithmic}	
\usepackage{setspace}           

\usepackage{epstopdf}	
\usepackage{enumerate} 
\usepackage{hyperref}
\usepackage{multicol} 
\usepackage{enumitem} 
\usepackage{cite} 
\usepackage{appendix}

\allowdisplaybreaks

\newcommand{\mypar}[1]{\bigskip\noindent {\bf #1.}}

\definecolor{red}{RGB}{153,0,0}

\title{Multimodal Image Denoising based on Coupled Dictionary Learning}
\name{
	Pingfan Song$^{\star}$ \quad
	Miguel R.\ D.\ Rodrigues$^{\star}$  
	\thanks{
		This work was supported by the Royal Society International Exchange Scheme IE160348, by UCL Overseas Research Scholarship (UCL-ORS) and by China Scholarship Council (CSC).
	}
}
\address{				
	$^{\star}$ Department of Electronic and Electrical Engineering, University College London, UK%
}

\begin{document}
\ninept
\maketitle

\begin{abstract}
	In this paper, we propose a new multimodal image denoising approach to attenuate white Gaussian additive noise in a given image modality under the aid of a guidance image modality. 
	%
	%
	The proposed coupled image denoising approach consists of two stages: coupled sparse coding and reconstruction. The first stage performs joint sparse transform for multimodal images with respect to a group of learned coupled dictionaries, followed by a shrinkage operation on the sparse representations. Then, in the second stage, the shrunken representations, together with coupled dictionaries, contribute to the reconstruction of the denoised image via an inverse transform.
	%
	%
	The proposed denoising scheme demonstrates the capability to capture both the common and distinct features of different data modalities. This capability makes our approach more robust to inconsistencies between the guidance and the target images,  thereby overcoming drawbacks such as the texture copying artifacts. Experiments on real multimodal images demonstrate that the proposed approach is able to better employ guidance information to bring notable benefits in the image denoising task with respect to the state-of-the-art.


\end{abstract}


\section{Introduction}
\label{sec:intro}

\vspace{-0.2cm}

Image Denoising is a type of techniques that attenuate the noise in the corrupted images and at the same time preserve the details faithfully. Serving as a fundamental image processing operation, image denoising plays a critical role in various application scenarios such as object detection, image recognition and remote sensing\cite{shao2014heuristic}.
Typical image denoising approaches that focus on single modality images have been thoroughly investigated, which can be categorized into two classes according to the image representation: spatial domain based local and nonlocal filters\cite{zhu2010automatic,bouboulis2010adaptive, buades2005non,nguyen2017bounded,talebi2014global,romano2015boosting}; and (predefined or learned) transform domain based approaches~\cite{dabov2007image, zhang2010two, chatterjee2012patch, elad2006image, mairal2009non}.

However, in many practical application scenarios, it is commonly noticed that a certain scene is often imaged using various sensors that yield different image modalities. For example, in remote sensing domain, it is typical to have various image modalities of earth observations, such as a panchromatic band version, a multispectral bands version, and an infrared (IR) band version~\cite{gomez2015multimodal,loncan2015hyperspectral}. These different bands often exhibit similar textures, edges, corners, boundaries, or other salient features. 
In colorization\cite{levin2004colorization} task, the output image has both chrominance channels and luminance channel which share consistent edges. These scenarios call for approaches that can capitalize on the availability of multiple image modalities of the same scene to address the image denoising task.
%
A variety of joint image filtering approaches have been proposed to capitalize on the availability of additional \emph{guidance} images, also referred to as \emph{side information}\cite{renna2016classification,mota2017compressed}, to aid the processing of target modalities\cite{kopf2007joint,he2013guided,ham2017robust,li2016deep,shen2015multispectral,zhang2014rolling}. 
The basic idea behind these methods is that the structural details of the guidance image can be transferred to the target image. However, these methods tend to introduce notable texture-copying artifacts, i.e. erroneous structure details that are not originally present in the target image because such methods typically fail to distinguish similarities and disparities between the different image modalities.

\begin{figure}[t]
	\centering
	\includegraphics[width= 9cm]{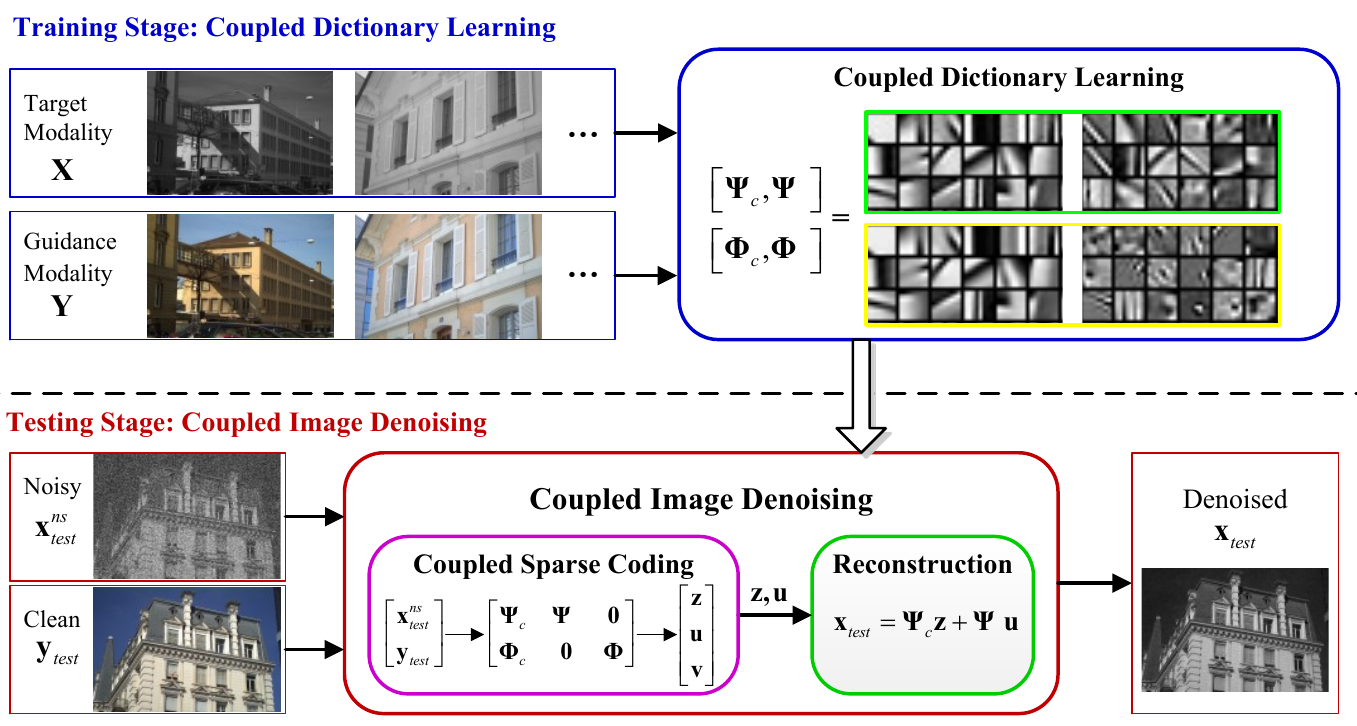}  
	\caption{Proposed coupled image denoising scheme.}
	\label{Fig:Diagram}
\end{figure}

This paper proposes a new effective multimodal image denoising approach based on coupled dictionary learning. With joint sparse representation induced by coupled dictionaries as the bridge, our approach incorporates a given clean guidance image as the side information to aid the denoising of the target image of interested modality. In particular, the proposed design has the ability to take into account both similarities and disparities between target and guidance images in order to deliver superior denoising performance. Further, we also incorporate self-similarity prior in our algorithm to enhance the denoising performance.

\section{Coupled Image Denoising}
\label{sec:SIMIS}

\vspace{-0.2cm}

In this section, We introduce the proposed multimodal data model, the coupled image denoising and the coupled dictionary learning strategy.
%


\vspace{-0.2cm}

\subsection{Multi-modal Data Model}


Given a vectorized noisy image $\mathbf{X}^{ns} \in \mathbb{R}^{N}$ of one modality and a corresponding registered clean vectorized guidance image $\mathbf{Y} \in \mathbb{R}^{N}$ of different modality as side information. We first extract (overlapping) image patch pairs from this pair of multimodal images. In particular, let $\mathbf{x}^{ns}_{i} = \mathbf{R}_i \mathbf{X}^{ns} \in \mathbb{R}^{n}$ denote the $i$-th noisy image patch from $\mathbf{X}^{ns}$ and let $\mathbf{y}_{i} = \mathbf{R}_i \mathbf{Y} \in \mathbb{R}^{n}$ denote the corresponding $i$-th clean guidance image patch extracted from $\mathbf{Y}$, where the matrix $\mathbf{R}_i$ is an $n \times N$ binary matrix that extracts the $i$-th patch from the image. Then, we propose a data model to capture the relationship -- including similarities and disparities -- between the two different modalities as follows:
\begin{align}
\mathbf{x}^{ns}_i 
&= 
\boldsymbol{\Psi}_{c} \, \mathbf{z}_i + \boldsymbol{\Psi} \, \mathbf{u}_i + \epsilon \,,
\label{Eq:SparseModelX_Noise}
\\
\mathbf{y}_i
&= 
\boldsymbol{\Phi}_{c} \, \mathbf{z}_i + \boldsymbol{\Phi} \, \mathbf{v}_i \,,
\label{Eq:SparseModelY_Noise}
\end{align}
where sparse representation $\mathbf{z}_i \in \mathbb{R}^{k}$ is common to both modalities, $\mathbf{u}_i \in \mathbb{R}^{k}$ is specific to modality $\mathbf{x}$, while $\mathbf{v}_i \in \mathbb{R}^{k}$ is specific to modality $\mathbf{y}$. In turn, $\boldsymbol{\Psi}_{c} $ and $\boldsymbol{\Phi}_{c} \in \mathbb{R}^{n \times k}$ are a pair of dictionaries associated with the common sparse representation $\mathbf{z}_i$, whereas $\boldsymbol{\Psi}$ and $\boldsymbol{\Phi} \in \mathbb{R}^{n \times k}$ are dictionaries associated with the specific sparse representations $\mathbf{u}_i$ and $\mathbf{v}_i$, respectively. 
Note that the common sparse representation $\mathbf{z}_i$ connects the patches of the two different modalities. The disparities between modalities $\mathbf{x}$ and $\mathbf{y}$ are distinguished by the sparse representations $\mathbf{u}_i$ and $\mathbf{v}_i$, respectively. 
Parameter $\epsilon \in \mathbb{R}^n$ denotes additive zero-mean and homogeneous white Gaussian noise with the standard deviation $\sigma$.


\vspace{-0.2cm}

\subsection{Coupled Image Denoising}
Assume that the coupled dictionaries $\boldsymbol{\Psi}_{c} $, $\boldsymbol{\Phi}_{c}, \boldsymbol{\Psi}, \boldsymbol{\Phi}$ have been learned, then, based on the model \eqref{Eq:SparseModelX_Noise} and \eqref{Eq:SparseModelY_Noise}, our coupled image denoising problem is addressed in two steps: coupled sparse coding and reconstruction of the denoised image.
\begin{algorithm}[t] 
	\caption{Coupled Dictionary Learning}
	\label{Alg:CoupledBCD}
		\begin{algorithmic}[1]	
			\renewcommand{\algorithmicrequire}{\textbf{Input:}}
			\renewcommand{\algorithmicensure}{\textbf{Output:}}
			\REQUIRE 
			Dataset 
			$\{(\mathbf{x}_i, \mathbf{y}_i) \}_{i=1}^T$, initial dictionaries $\boldsymbol{\Psi}_{c}, \boldsymbol{\Phi}_{c},\boldsymbol{\Psi}, \boldsymbol{\Phi} \in \mathbb{R}^{n \times k}$ (randomly selected patches), regularization parameter $\lambda$.
			
			
			\renewcommand{\algorithmicrequire}{\textbf{Optimization:}}
			\REQUIRE		
			\STATE
			\textbf{Common dictionary training.} Iterating between a) and b).
			\STATE
			a) Update the sparse codes $\mathbf{Z}$ as :
			\begin{equation*}
			\mathbf{z}^j \leftarrow S_\lambda\left( \mathbf{z}^j + \frac{1}{\|\mathbf{d}_j \|_2^2} \mathbf{d}_j^T (\widetilde{\mathbf{X}} - \mathbf{D} \mathbf{Z}) \right) ; \forall j=1,\cdots, k
			\end{equation*}
			where 
			$\mathbf{z}^j$ denotes the $j$-th row of $\mathbf{Z}$, $\mathbf{d}_j$ denotes the $j$-th atom of dictionary $\mathbf{D}	= 
			\begin{bmatrix}
			\boldsymbol{\Psi}_{c}  \\
			\boldsymbol{\Phi}_{c}  \\
			\end{bmatrix} $,
			$\widetilde{\mathbf{X}} = 
			\begin{bmatrix} 
			\mathbf{X} - \boldsymbol{\Psi} \mathbf{U} \\ 
			\mathbf{Y} -  \boldsymbol{\Phi} \mathbf{V}
			\end{bmatrix} $.
			$S_\lambda(\cdot)$ denotes the element-wise soft-thresholding operator
			$S_\lambda(\alpha) = \text{sign}(\alpha) \max(|\alpha|-\lambda, 0)$.
			
			\STATE
			b) Update the dictionary $\boldsymbol{\Psi}_{c}$ and $\boldsymbol{\Phi}_{c}$ as :
			\begin{align*}
			\mathbf{d}_j &\leftarrow \frac{1}{\mathbf{z}^j {\mathbf{z}^j}^T} (\widetilde{\mathbf{X}} - \mathbf{D}\mathbf{Z}) {\mathbf{z}^j}^T +
			\begin{bmatrix}
			\boldsymbol{\psi}_{cj}  \\
			\boldsymbol{\phi}_{cj}  \\
			\end{bmatrix}
			\\
			\begin{bmatrix}
			\boldsymbol{\psi}_{cj}  \\
			\boldsymbol{\phi}_{cj}  \\
			\end{bmatrix}
			&\leftarrow \frac{\mathbf{d}_j}{\max(\|\mathbf{d}_j \|_2, 1)} 
			; \forall j=1,\cdots, k.
			\end{align*}
			
			\STATE
			\textbf{Unique dictionary training.} Iterating between a) and b).
			\STATE
			a) Update the sparse codes $\mathbf{U}$ as :
			\begin{equation*}
			\mathbf{u}^j \leftarrow S_\lambda\left( \mathbf{u}^j + \frac{1}{\|\boldsymbol{\psi}_j \|_2^2} \boldsymbol{\psi}_j^T (\mathbf{X} - \boldsymbol{\Psi}_c \mathbf{Z} - \boldsymbol{\Psi} \mathbf{U}) \right) ; \forall j
			\end{equation*}
			where $\mathbf{u}^j$ denotes the $j$-th row of $\mathbf{U}$, $\boldsymbol{\psi}_j$ denotes the $j$-th atom of dictionary $\boldsymbol{\Psi}$,
			
			\STATE
			b) Update the dictionary $\boldsymbol{\Psi}$ as :
			\begin{align*}
			\boldsymbol{\psi}_j &\leftarrow \frac{1}{\mathbf{u}^j {\mathbf{u}^j}^T} (\mathbf{X} - \boldsymbol{\Psi}_c \mathbf{Z} - \boldsymbol{\Psi} \mathbf{U}) {\mathbf{u}^j}^T + \boldsymbol{\psi}_j 
			\\
			\boldsymbol{\psi}_j
			&\leftarrow \frac{\boldsymbol{\psi}_j}{\max(\|\boldsymbol{\psi}_j \|_2, 1)} 
			; \forall j=1,\cdots, k.
			\end{align*}
			
			\STATE
			\textrm{Unique dictionary training for $\boldsymbol{\Phi}$ is similar to $\boldsymbol{\Psi}$ but with $\mathbf{u}^j$, $\boldsymbol{\psi}_j$, $\mathbf{X}$, $\boldsymbol{\Psi}_c$ replaced by $\mathbf{v}^j$, $\boldsymbol{\phi}_j$}, $\mathbf{Y}$, $\boldsymbol{\Phi}_c$.
			%
			\STATE
			\textbf{Return} dictionaries $\boldsymbol{\Psi}_{c}, \boldsymbol{\Phi}_{c},\boldsymbol{\Psi}, \boldsymbol{\Phi}$.
		\end{algorithmic}
\end{algorithm}	

\mypar{Step 1: Coupled sparse coding}
In this stage, we address a series of parallel sparse coding problems formulated as
%
\begin{equation} \label{Eq:BPDN}
\small 
\begin{array}{cl}
\underset{\mathbf{z}_i, \mathbf{u}_i,\mathbf{v}_i}{\min} 
\,
\left\| 
\begin{bmatrix}
\mathbf{z}_i \\
\mathbf{u}_i \\
\mathbf{v}_i \\
\end{bmatrix} 
\right\|_0
\,
\textrm{s.t. }
\;
\left\|
\begin{bmatrix} 
\mathbf{x}^{ns}_{i} \\ \mathbf{y}_{i}
\end{bmatrix}
-
\begin{bmatrix}
\boldsymbol{\Psi}_{c} & \boldsymbol{\Psi} & \mathbf{0} \\
\boldsymbol{\Phi}_{c} & \mathbf{0} & \boldsymbol{\Phi} \\
\end{bmatrix}
\begin{bmatrix}
\mathbf{z}_i \\
\mathbf{u}_i \\
\mathbf{v}_i \\
\end{bmatrix}
\right\|_2^2
\leq C \, \sigma^2
\end{array}
\end{equation}
where the standard deviation $\sigma$ represents the noise level, and $C$ is a constant. The objective with $\ell_0$-penalty serves as the sparsity-inducing regularizer.\footnote{
	We denote by $\| x\|_0$ the number of nonzero elements of the vector $x$. Note, this $\ell_0$ sparsity measure is a pseudo norm as it does not preserve the homogeneity property. 
}
The quadratic "data-fitting" constraint ensures that each pair of multimodal image patches are well approximated by their sparse representations with respect to the coupled dictionaries. 
In our case, we stick to $\ell_0$ penalty for the sparse coding as it usually leads to better denoising performance than $\ell_1$ penalty, which is also observed in \cite{mairal2009non,mairal2014sparse}.

\vspace{-0.2cm}

\mypar{Step 2: Reconstruction}
Given the sparse codes $\mathbf{z}_i,\mathbf{u}_i$, the denoised image of interested modality can be estimated by solving \eqref{Eq:Denoise_Update}, which leads to a closed form solution \eqref{Eq:Denoise_Update2}.
\begin{equation} \label{Eq:Denoise_Update}
\footnotesize
\underset{\mathbf{X}_{} }{\min} \;
\mu \left\| \mathbf{X}_{} - \mathbf{X}^{ns}_{} \right\|_2^2
+
\sum_i
\left\| \mathbf{R}_i \mathbf{X} - ( \boldsymbol{\Psi}_{c} \mathbf{z}_i + \boldsymbol{\Psi} \mathbf{u}_i) \right\|_2^2
\end{equation}
\begin{equation} \label{Eq:Denoise_Update2}
\footnotesize
\mathbf{\widehat{X}} = \Big( \mu \mathbf{I} + \sum_{i} \mathbf{R}_i^T \mathbf{R}_i \Big)^{-1} 
\Big( \mu \mathbf{X}^{ns} + \sum_i \mathbf{R}_i^T (\boldsymbol{\Psi}_{c} \mathbf{z}_i + \boldsymbol{\Psi} \mathbf{u}_i) \Big) 
\end{equation} 
For $\mu = 0$, this expression just represents the average of the denoised image patches on the overlapping areas, leading to a purified image. For non-zero $\mu$, the estimation operation also refers to the original noisy image during the average.

\subsection{Coupled Image Denoising with Group Sparsity}
It is widely observed that natural images exhibit significant self-similarity property. The self-similarity prior is exploited to average out the noise among multiple similar patches\cite{dabov2007image,mairal2009non}. Similar phenomena are also commonly observed in multimodal images, which motivates us to integrate self-similarity prior into our scheme via group sparsity regularization, which imposes that similar patch pairs share the same sparsity patterns in their representations.

The grouped-sparsity of a matrix $\mathbf{A} \in \mathbb{R}^{3k \times l}$ is defined as
$$
\|\mathbf{A} \|_{p,q} := 
\left\|
\begin{bmatrix}
\mathbf{z}_1 & \ldots & \mathbf{z}_l \\
\mathbf{u}_1 & \ldots & \mathbf{u}_l \\
\mathbf{v}_1 & \ldots & \mathbf{v}_l \\
\end{bmatrix}
\right\|_{p,q}
=
\sum_{i=1}^{k} 
\| \mathbf{z}^i \|_q^p + \| \mathbf{u}^i \|_q^p + \| \mathbf{v}^i \|_q^p
$$
where, $\mathbf{z}^i$, $\mathbf{u}^i$, $\mathbf{v}^i$ denote the $i$-th row of matrix $[\mathbf{z}_1, \ldots, \mathbf{z}_l]$, $[\mathbf{u}_1, \ldots, \mathbf{u}_l]$, and $[\mathbf{v}_1, \ldots, \mathbf{v}_l]$, respectively. In our case, we choose $(p,q) = (0,\infty)$, leading to $\ell_{0,\infty}$ matrix pseudo norm, as a generalization of $\ell_0$ vector pseudo norm. This amounts to restricting the number of nonzero rows of $\mathbf{A}$. 

To obtain groups of similar patches, we apply hierarchical clustering to separate all the patch pairs into $M$ different clusters according to their similarity. 
Given the learned coupled dictionaries, sparse coding for the $i$-th cluster $S_i$ of patch pairs with a grouped-sparsity regularizer amounts to solving
\begin{equation} \label{Eq:BPDN_GroupSparsity}
\begin{array}{cl}
\underset{\mathbf{A}_i}{\min}
& \! \! \! \!
\| \mathbf{A}_i \|_{p,q} \;
\\
\text{s.t.} 
& \! \! \! \!
\sum\limits_{j \in S_i} 
\left\|
\begin{bmatrix} 
\mathbf{x}^{ns}_{j} \\ \mathbf{y}_{j}
\end{bmatrix}
-
\begin{bmatrix}
\boldsymbol{\Psi}_{c} & \boldsymbol{\Psi} & \mathbf{0} \\
\boldsymbol{\Phi}_{c} & \mathbf{0} & \boldsymbol{\Phi} \\
\end{bmatrix}
\begin{bmatrix}
\mathbf{z}_j \\
\mathbf{u}_j \\
\mathbf{v}_j \\
\end{bmatrix}
\right\|_2^2
\leq
|S_i| C \sigma^2  
\end{array}
\end{equation}
where $|S_i|$ denotes the cardinality of the cluster $S_i$.

\vspace{-0.2cm}

\subsection{Coupled Dictionary Learning}


The coupled image denoising require a group of coupled dictionaries to perform the sparse transform/decomposition task. 
These required coupled dictionaries are learned using our couple dictionary learning algorithm.

Given a corpus of registered image patch pairs $\{(\mathbf{x}_i, \mathbf{y}_i) \}_{i=1}^T$, our coupled dictionary learning problem is posed as follows:
%
\begin{equation} \label{Eq:CoupledDL_Clean}
\small
\begin{array}{cl}
\underset{ 
	\begin{subarray}{c}
	\left\{ \boldsymbol{\Psi}_{c}, \boldsymbol{\Psi}, \boldsymbol{\Phi}_{c}, \boldsymbol{\Phi} \right\}  \\
	\{ \mathbf{z}_i, \mathbf{u}_i, \mathbf{v}_i \}
	\end{subarray}}
{ \text{minimize}}
&  \! \! \! \!
\sum\limits_{i=1}^T
\frac{1}{2}
\left\|
\begin{bmatrix}
\mathbf{x}_i \\ \mathbf{y}_i 
\end{bmatrix}
-
\begin{bmatrix}
\boldsymbol{\Psi}_{c} & \boldsymbol{\Psi} & \mathbf{0} \\
\boldsymbol{\Phi}_{c} & \mathbf{0} & \boldsymbol{\Phi} \\
\end{bmatrix}
\begin{bmatrix}
\mathbf{z}_i \\
\mathbf{u}_i \\
\mathbf{v}_i \\
\end{bmatrix}
\right\|_2^2
+
\lambda
\left\| 
\begin{bmatrix}
\mathbf{z}_i \\
\mathbf{u}_i \\
\mathbf{v}_i \\
\end{bmatrix} 
\right\|_1
\\
\text{subject to}
& \! \! \! \!
\left\| \begin{bmatrix} \psi_{cj} \\ \phi_{cj} \end{bmatrix} \right\|_2^2 \leq 1, 
\| \psi_j \|_2^2 \leq 1, \| \phi_j \|_2^2 \leq 1, \forall j.
\end{array}
\end{equation}
\noindent
where, the first term with $\ell_2$ norm promotes the fidelity of sparse representations to the signals and the regularization term with $\ell_1$ norm promotes sparsity for the representations. The $\ell_2$ norm constraints for each atom pair $[\psi_{cj} ; \phi_{cj}]$ and individual atom $\psi_j$ and $\phi_j$ are used to avoid trivial solution.

The coupled dictionary learning problems in \eqref{Eq:CoupledDL_Clean} is a non-convex optimization problem. 
We solve it via an alternating optimization scheme that performs a) sparse coding and b) dictionary update alternatively. During the sparse coding stage, we fix the all the dictionaries and obtain the sparse representations, while during the dictionary updating stage, we fix the sparse codes and update the all the dictionaries. The dictionary updating formulations are adapted from Block Coordinate Descent algorithm~\cite{mairal2010online}, where we train the common dictionaries simultaneously while train the unique dictionaries individually, as shown in Algorithm \ref{Alg:CoupledBCD}.

\begin{table*}[t]
	\scriptsize
	\centering
	\caption{Multimodal image denoising performance in terms of average PSNR and RMSE at different noise levels.}
	\begin{tabular}{l| ll| ll| ll| ll| ll| ll |ll }
		\hline \hline
		Noise & \multicolumn{2}{c|}{JBF\cite{kopf2007joint}} & \multicolumn{2}{c|}{GF\cite{he2013guided}} & \multicolumn{2}{c|}{SDF\cite{ham2017robust}} & \multicolumn{2}{c|}{DJF\cite{li2016deep}} & \multicolumn{2}{c|}{JFSM\cite{shen2015multispectral}} & \multicolumn{2}{c|}{Proposed} & \multicolumn{2}{c}{Proposed+} \\
		$\sigma$/PSNR & RMSE & PSNR & RMSE & PSNR & RMSE & PSNR & RMSE & PSNR & RMSE & PSNR & RMSE & PSNR & RMSE & PSNR \\
		\hline
		4/36.08 & 0.0153 & 36.75 & 0.0111 & 39.51 & 0.0163 & 35.97 & 0.0119 & 38.84 & 0.0170 & 35.45 & \textbf{0.0085} & \textbf{41.57} & 0.0090 & 41.07 \\
		8/30.07 & 0.0172 & 35.66 & 0.0152 & 36.78 & 0.0172 & 35.46 & 0.0137 & 37.46 & 0.0171 & 35.39 & \textbf{0.0116} & \textbf{38.88} & 0.0117 & 38.80 \\
		12/26.54 & 0.0189 & 34.82 & 0.0167 & 35.87 & 0.0186 & 34.76 & 0.0162 & 36.11 & 0.0184 & 34.76 & 0.0140 & 37.18 & \textbf{0.0139} & \textbf{37.28} \\
		16/24.05 & 0.0203 & 34.16 & 0.0185 & 35.01 & 0.0204 & 33.90 & 0.0180 & 35.19 & 0.0198 & 34.12 & 0.0162 & 35.93 & \textbf{0.0159} & \textbf{36.12} \\
		20/22.11 & 0.0215 & 33.63 & 0.0200 & 34.34 & 0.0227 & 32.95 & 0.0198 & 34.45 & 0.0216 & 33.38 & 0.0182 & 34.92 & \textbf{0.0178} & \textbf{35.13} \\
		24/20.52 & 0.0227 & 33.16 & 0.0213 & 33.79 & 0.0255 & 31.93 & 0.0209 & 33.91 & 0.0232 & 32.75 & 0.0201 & 34.03 & \textbf{0.0195} & \textbf{34.30} \\
		\hline \hline
	\end{tabular}
	\label{Tab:PSNR_RMSE}
\end{table*}

%
\begin{figure}[t]
	\centering
	\includegraphics[width = 7cm]{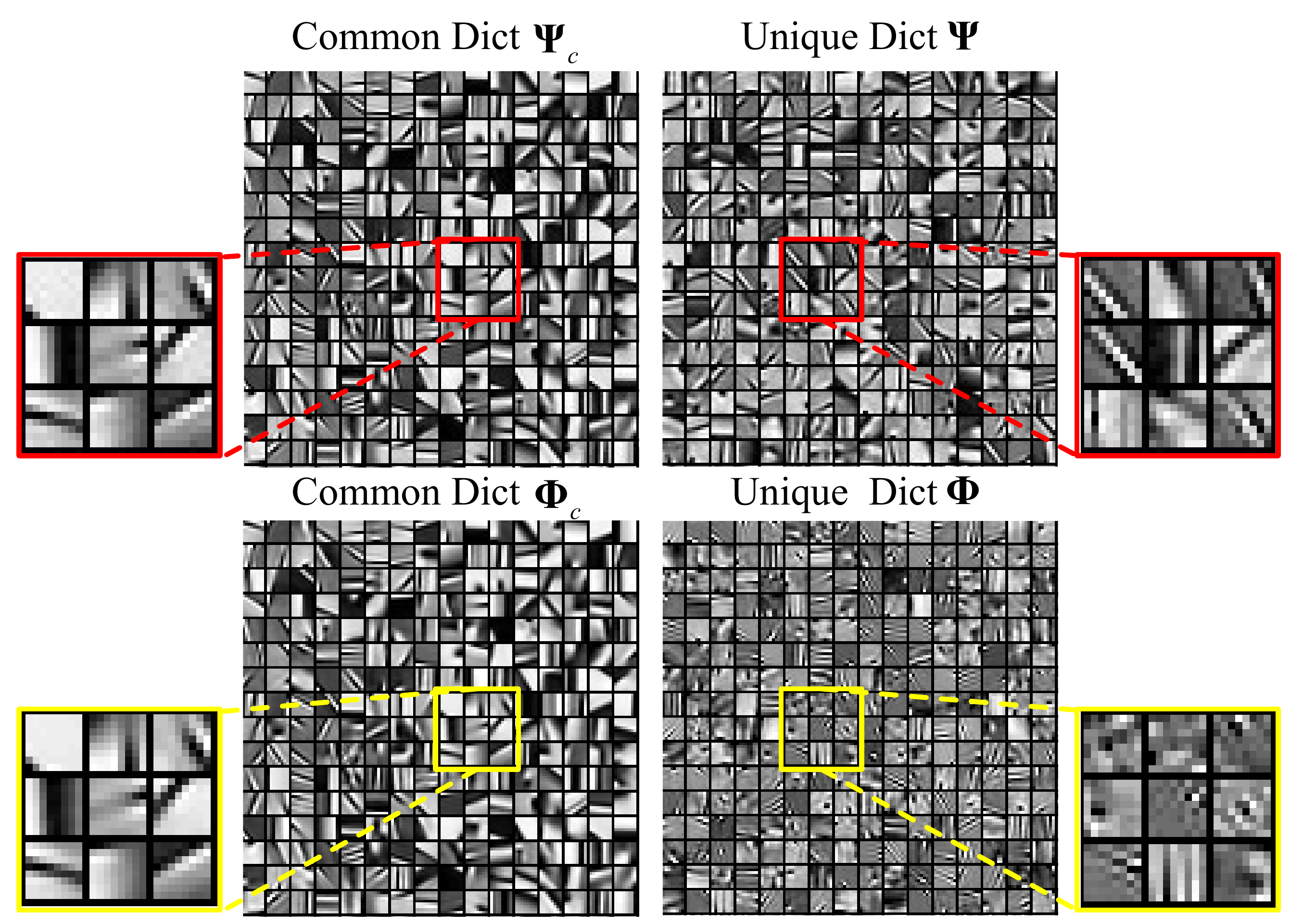} 
	
	\vspace{-0.2cm}
	
	\caption{Learned coupled dictionaries for infrared/RGB images. 256 atoms are shown here. The top two indicate the common and unique dictionaries learned for infrared images. The bottom two display dictionaries learned from corresponding guidance modality.}
	\label{Fig:LearnedD}
\end{figure}

\begin{figure}[t]
	\centering
	\begin{minipage}[b]{0.48\linewidth}
		\centering
		\includegraphics[width = 4.2cm, height = 2.8cm]{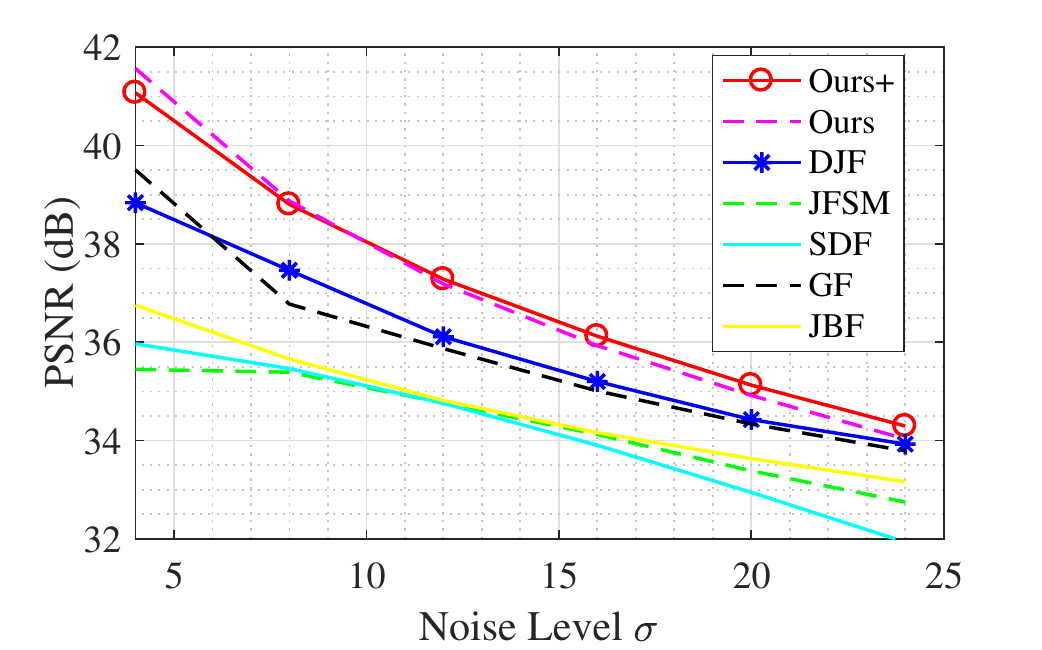} 
	\end{minipage}
	\begin{minipage}[b]{0.48\linewidth}
		\centering
		\includegraphics[width = 4.2cm, height = 2.8cm]{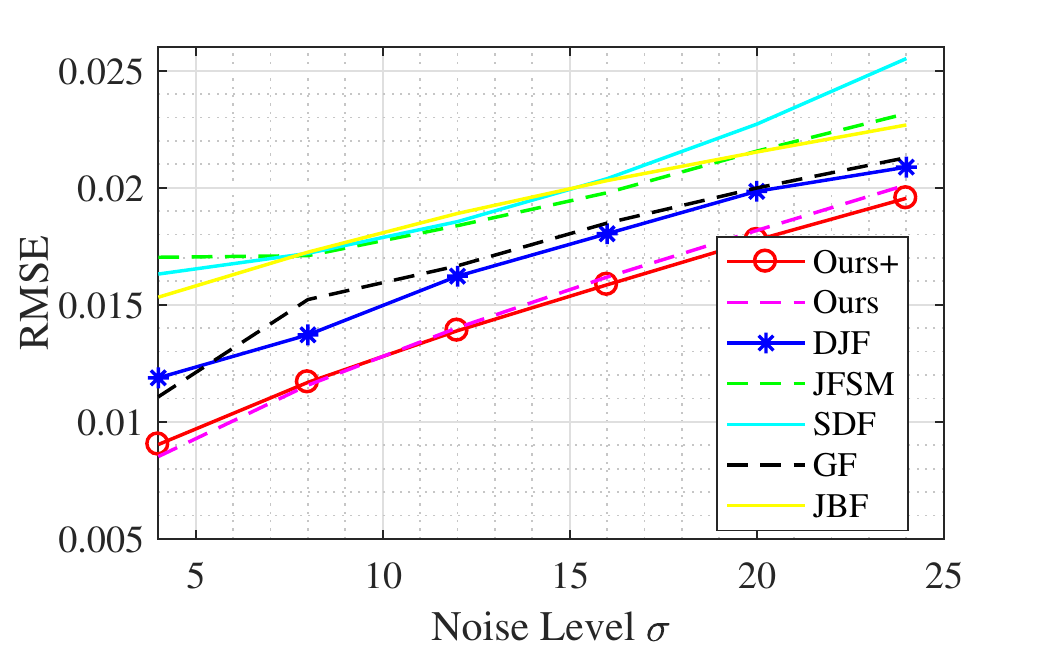}
	\end{minipage}

	\vspace{-0.2cm}
	
	\caption{Multimodal image denoising in terms of PSNR and RMSE with respect to noise level. We compare our basic approach (Ours) and the advanced version (Ours+) with state-of-the-art joint image filtering approaches, such as JBF\cite{kopf2007joint}, GF\cite{he2013guided}, SDF\cite{ham2017robust}, JFSM\cite{shen2015multispectral} and DJF\cite{li2016deep}.}
	\label{Fig:PSNR_RMSE}
\end{figure}

\begin{figure*}[th]
	\begin{multicols}{2}  
		\centering
		\begin{minipage}[b]{0.1\linewidth}
			\centering
			{\footnotesize Truth}  
		\end{minipage} 
		\begin{minipage}[b]{0.28\linewidth}
			\centering
			\includegraphics[width = 2.5cm, height = 2.5cm]{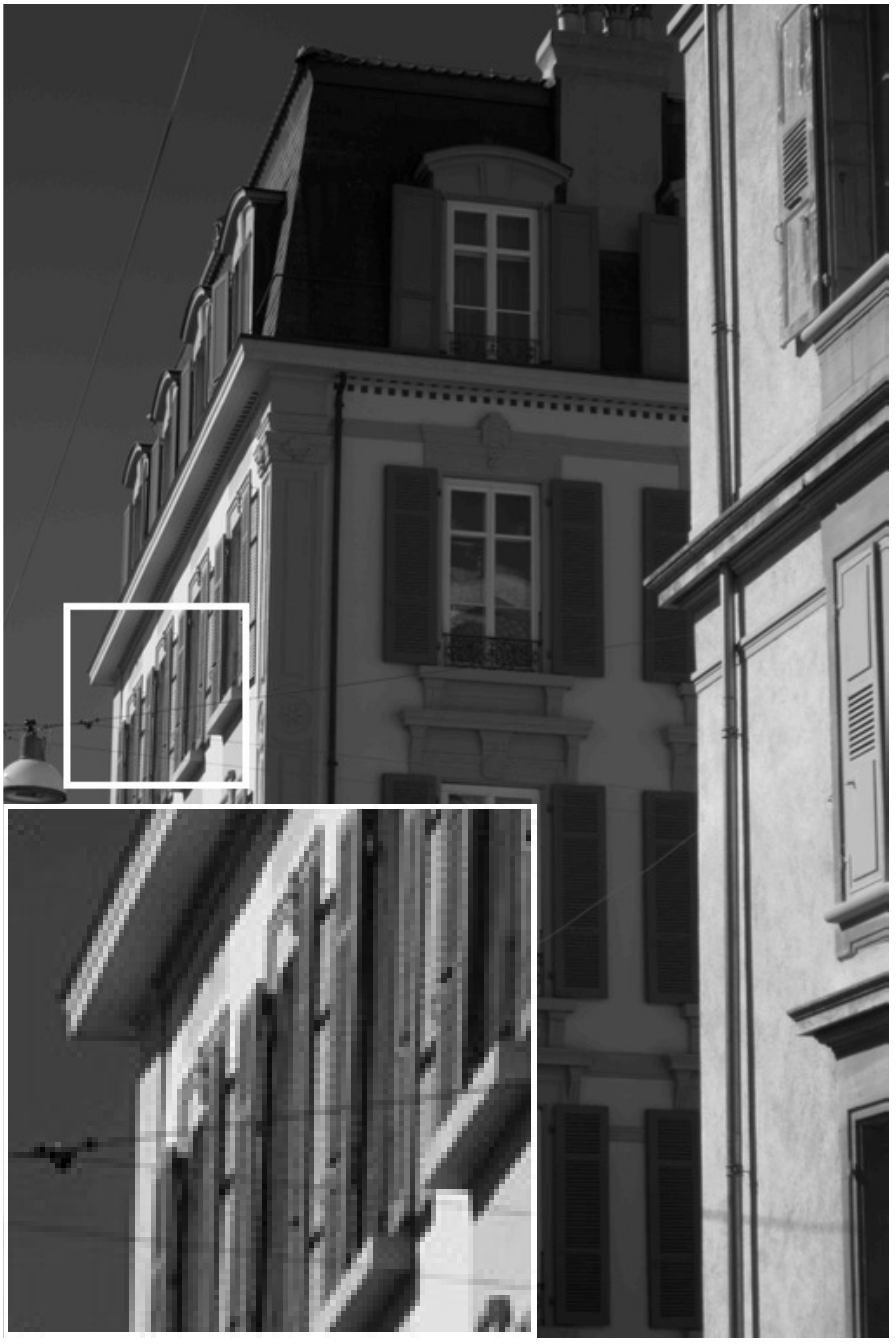}
		\end{minipage} 
		\begin{minipage}[b]{0.28\linewidth}
			\centering
			\includegraphics[width = 2.5cm, height = 2.5cm]{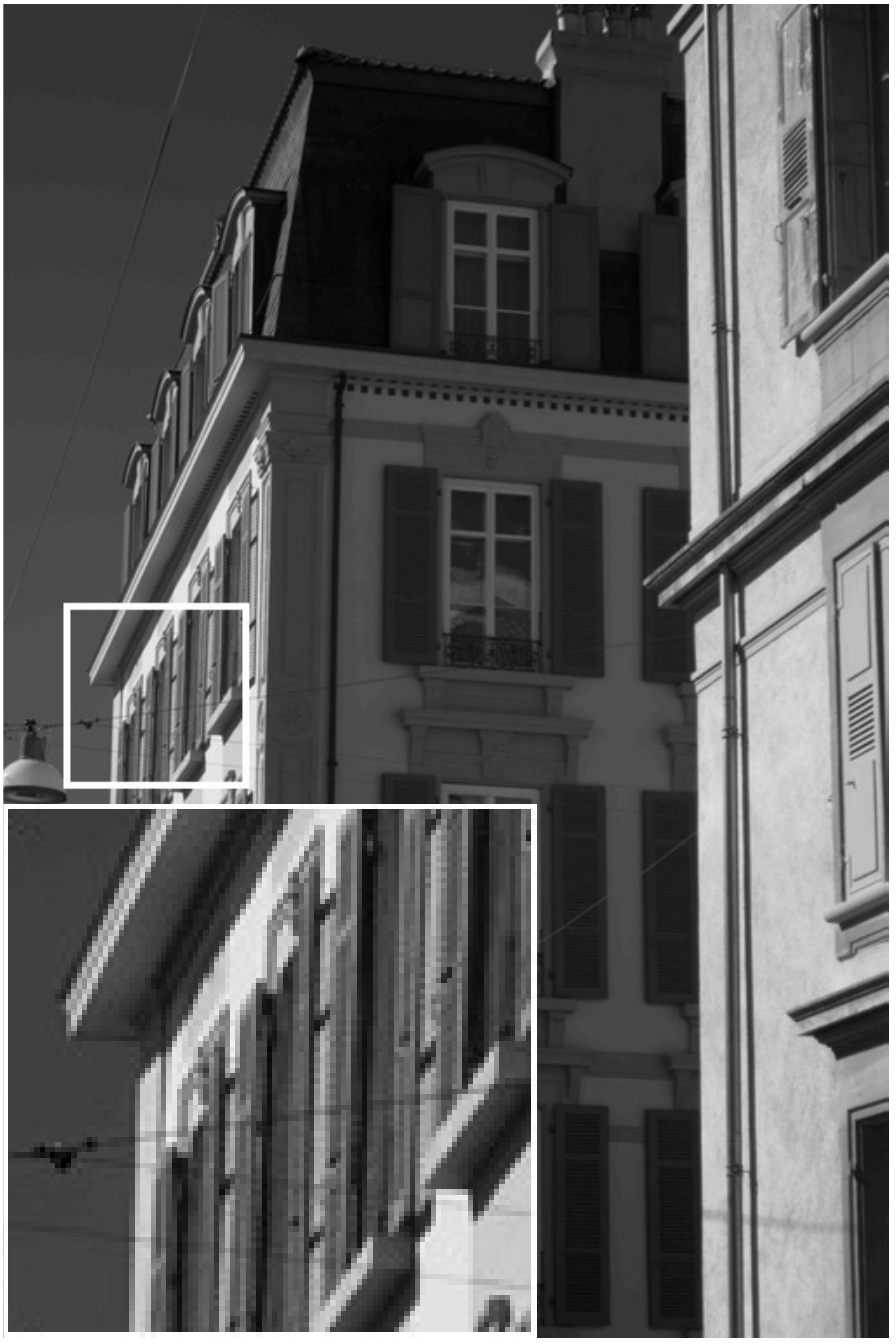}
		\end{minipage} 
		\begin{minipage}[b]{0.28\linewidth}
			\centering
			\includegraphics[width = 2.5cm, height = 2.5cm]{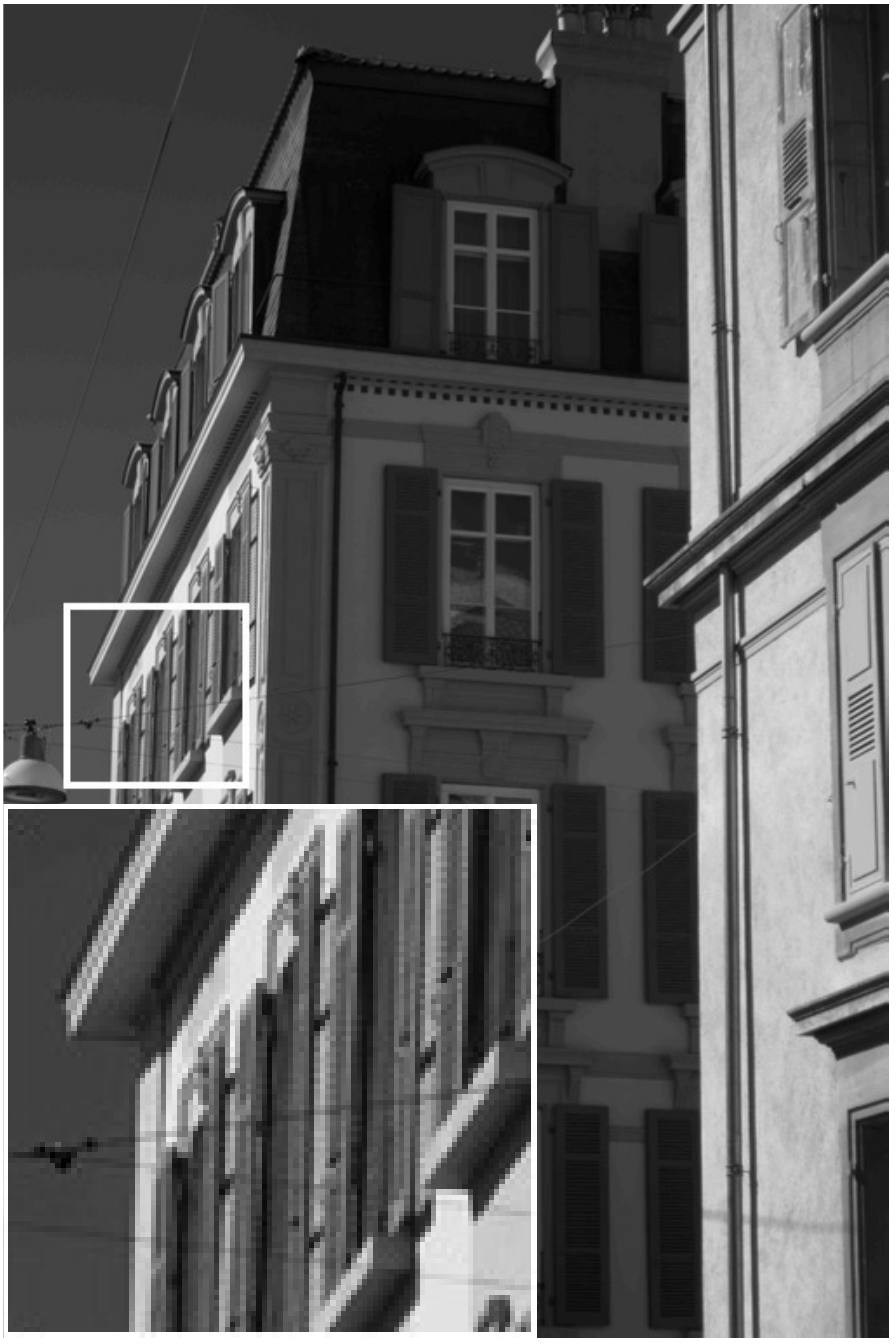}
		\end{minipage} 
		\\
		\begin{minipage}[b]{0.1\linewidth}
			\centering
			{\footnotesize Input}  
		\end{minipage} 
		\begin{minipage}[b]{0.28\linewidth}
			\centering
			\includegraphics[width = 2.5cm, height = 2.5cm]{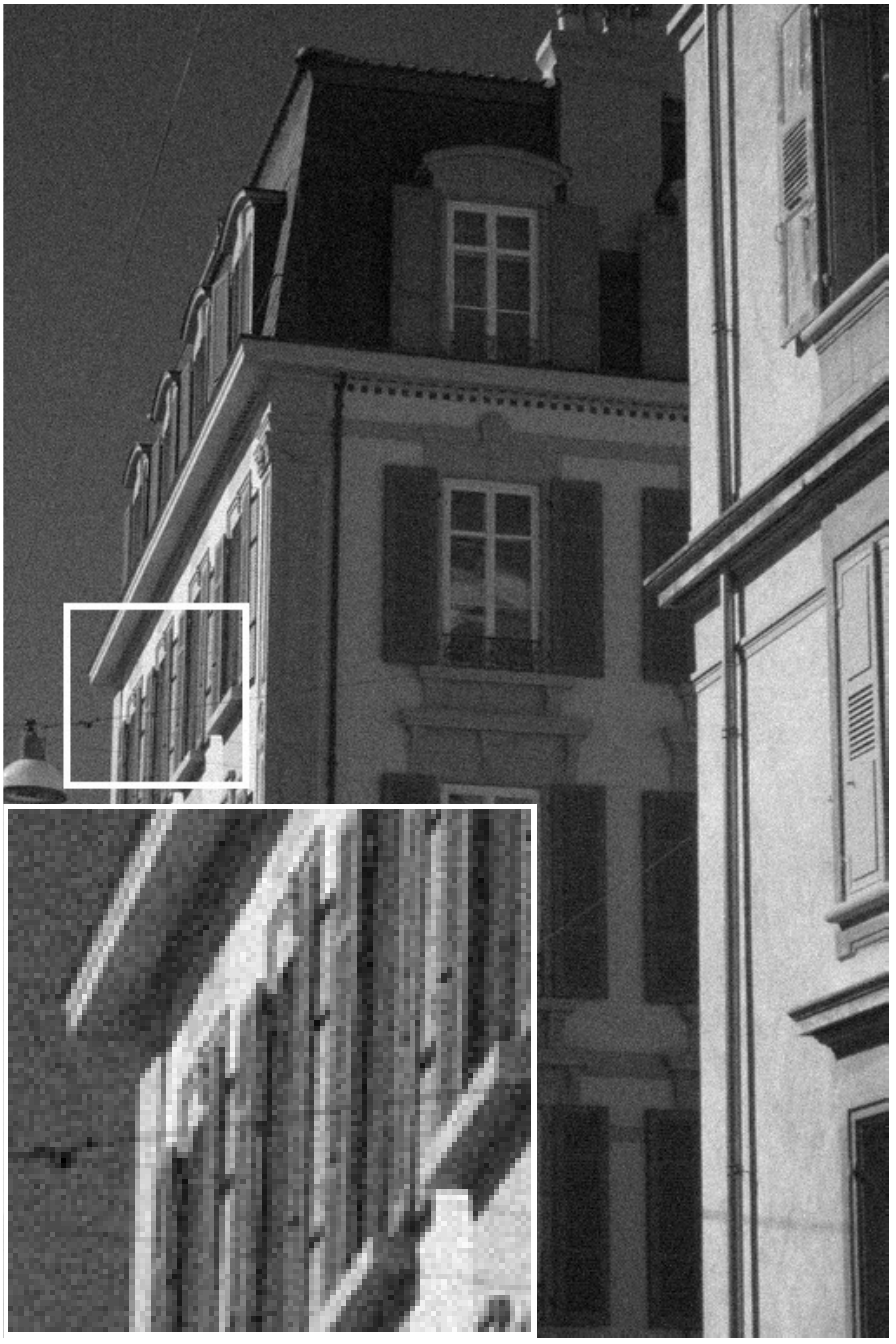}
		\end{minipage} 
		\begin{minipage}[b]{0.28\linewidth}
			\centering
			\includegraphics[width = 2.5cm, height = 2.5cm]{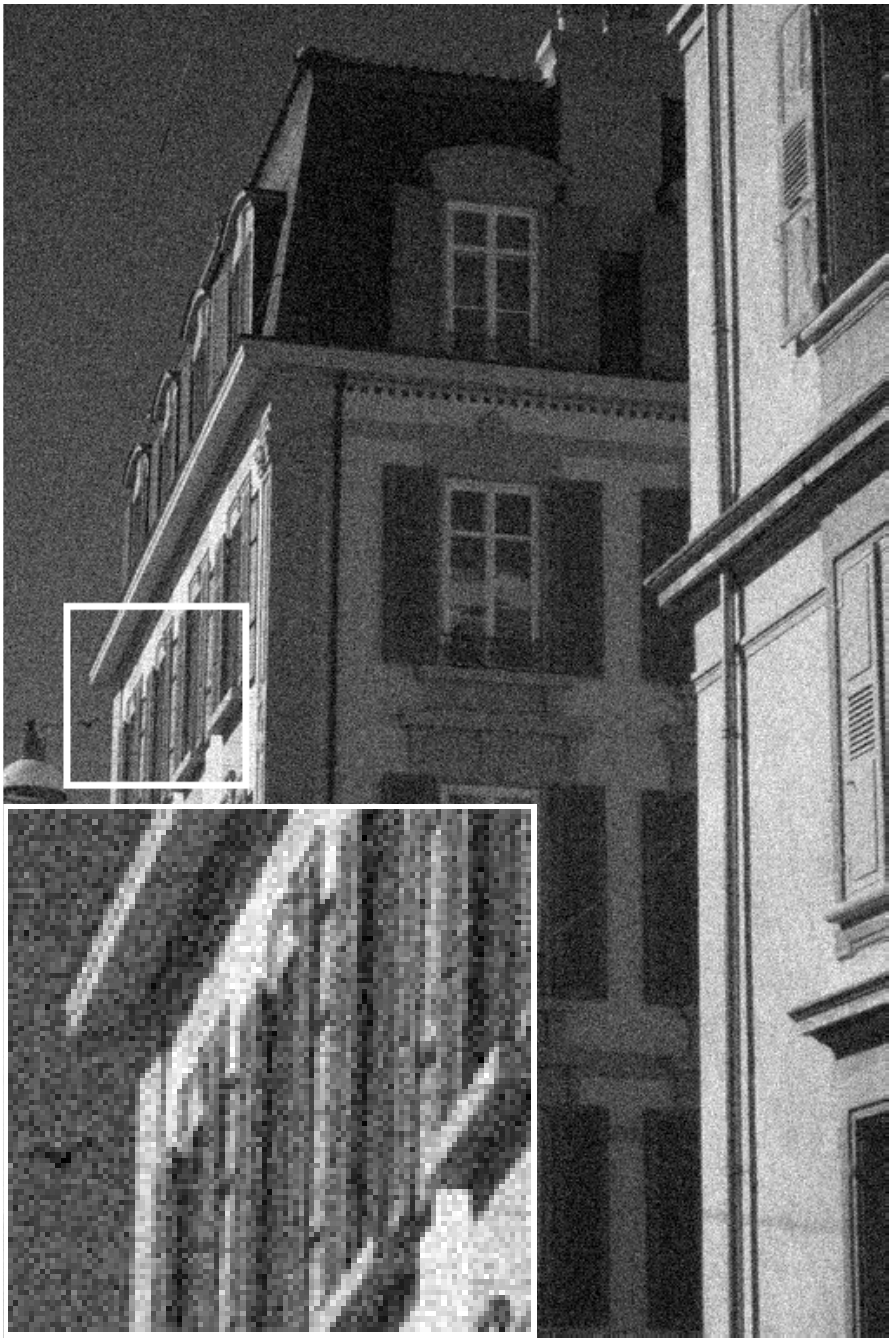}
		\end{minipage} 
		\begin{minipage}[b]{0.28\linewidth}
			\centering
			\includegraphics[width = 2.5cm, height = 2.5cm]{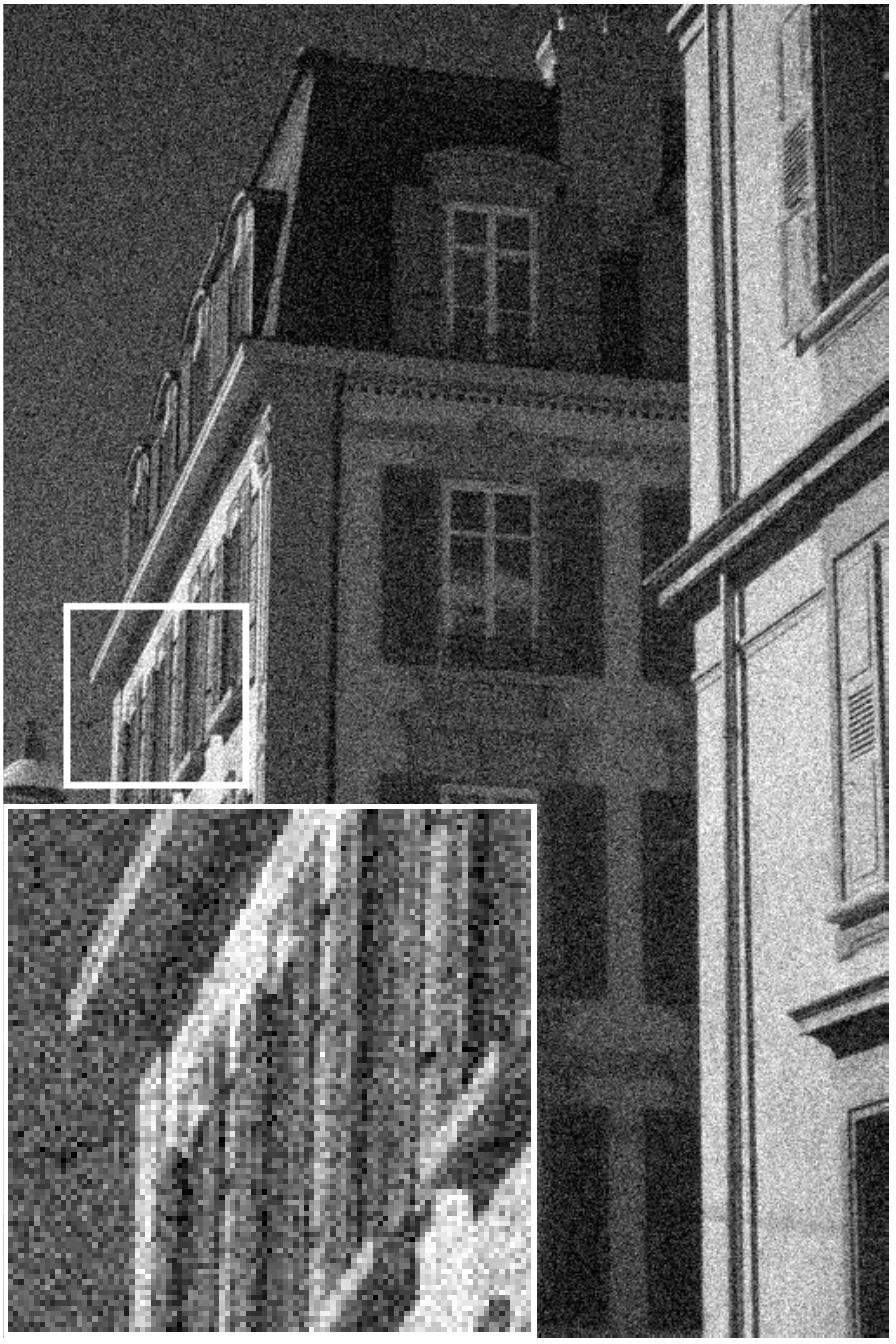}
		\end{minipage} 
		\\
		\begin{minipage}[b]{0.1\linewidth}
			\centering
			{\footnotesize GF~\cite{he2013guided}}  
		\end{minipage}  
		\begin{minipage}[b]{0.28\linewidth}
			\centering
			\includegraphics[width = 2.5cm, height = 2.5cm]{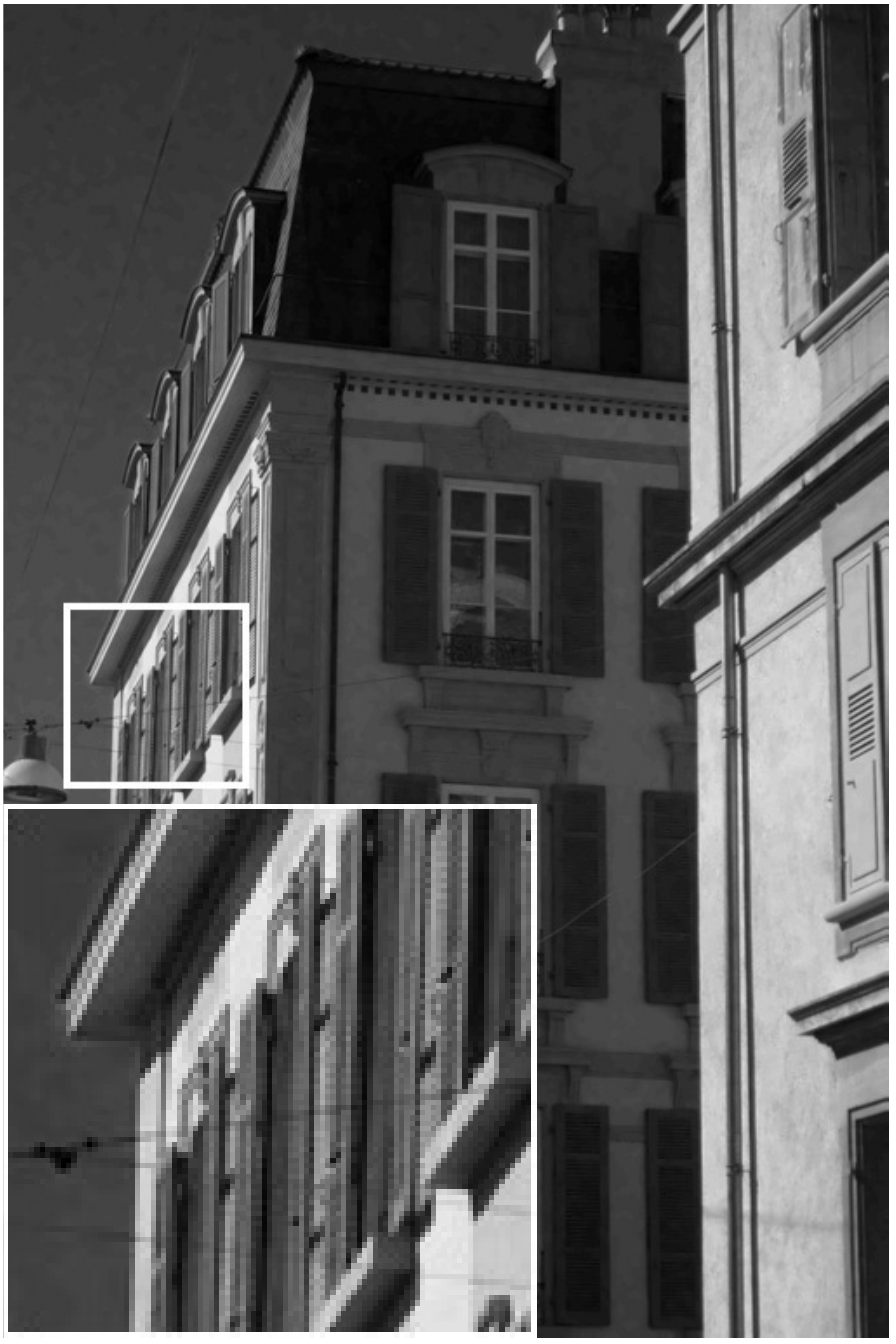}
		\end{minipage} 
		\begin{minipage}[b]{0.28\linewidth}
			\centering
			\includegraphics[width = 2.5cm, height = 2.5cm]{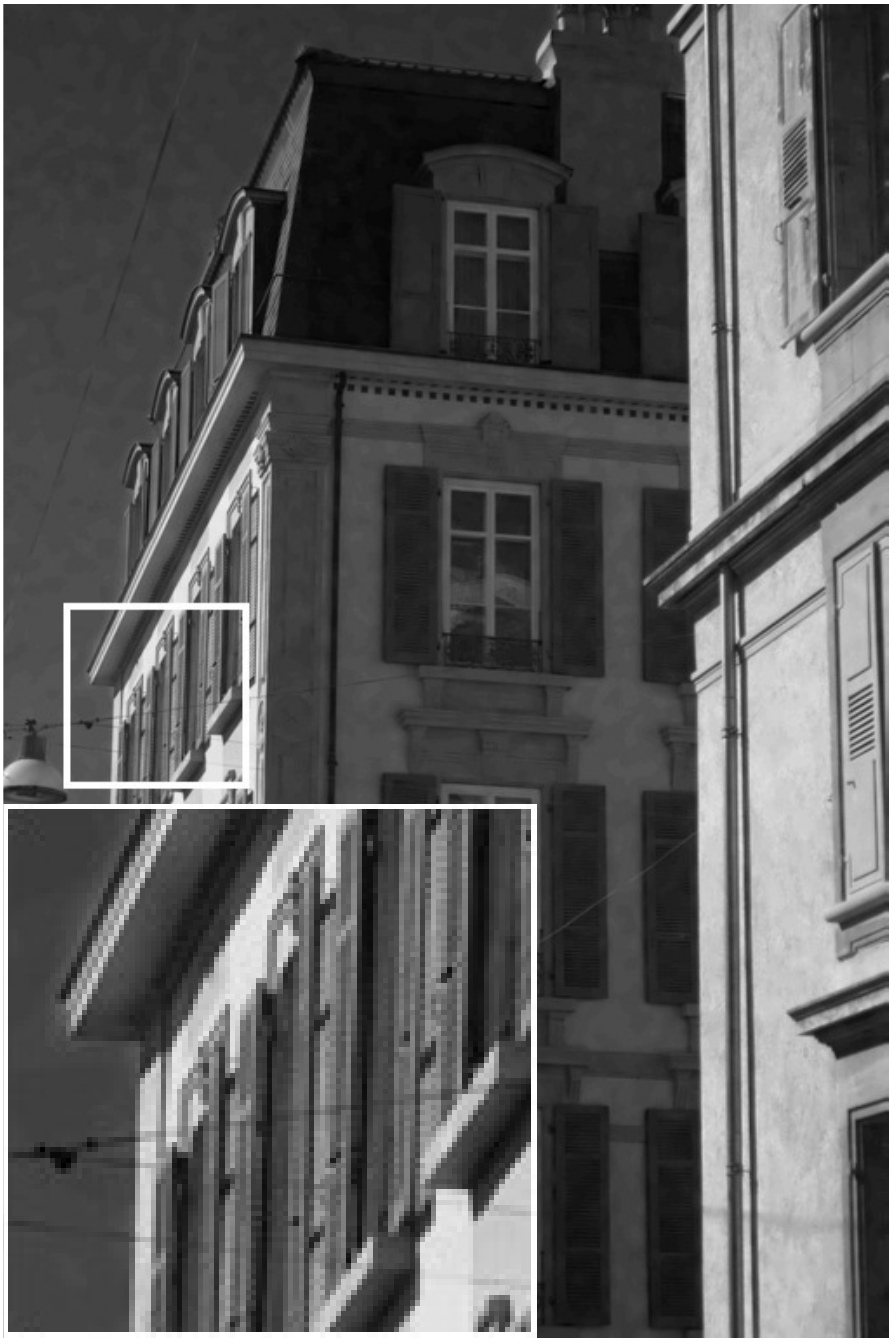}
		\end{minipage} 
		\begin{minipage}[b]{0.28\linewidth}
			\centering
			\includegraphics[width = 2.5cm, height = 2.5cm]{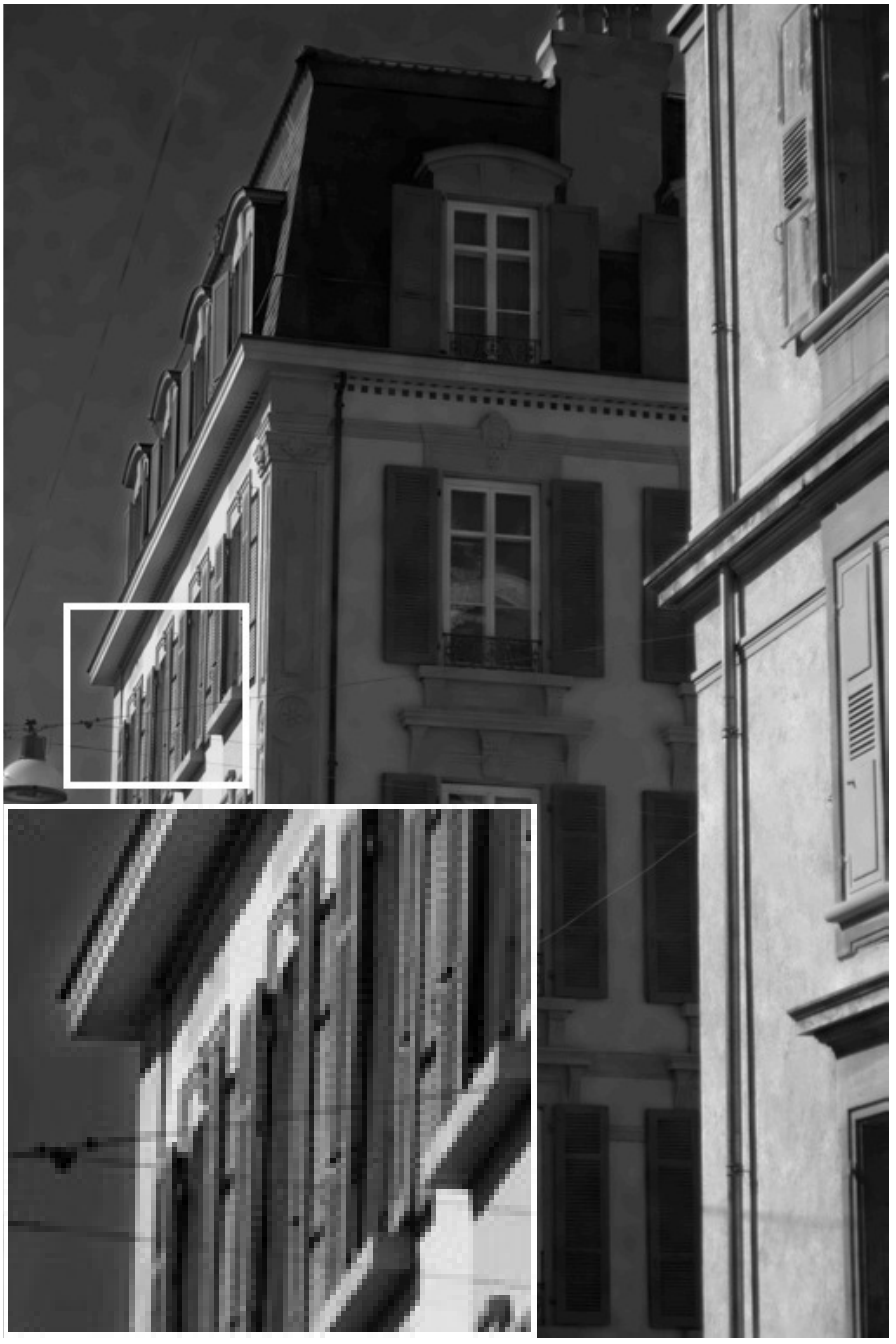}
		\end{minipage} 
		\\
		\begin{minipage}[b]{0.1\linewidth}
			\centering
			{\footnotesize ErrMap GF~\cite{he2013guided}}  
		\end{minipage} 
		\begin{minipage}[b]{0.28\linewidth}
			\centering
			\includegraphics[width = 2.5cm, height = 2.5cm]{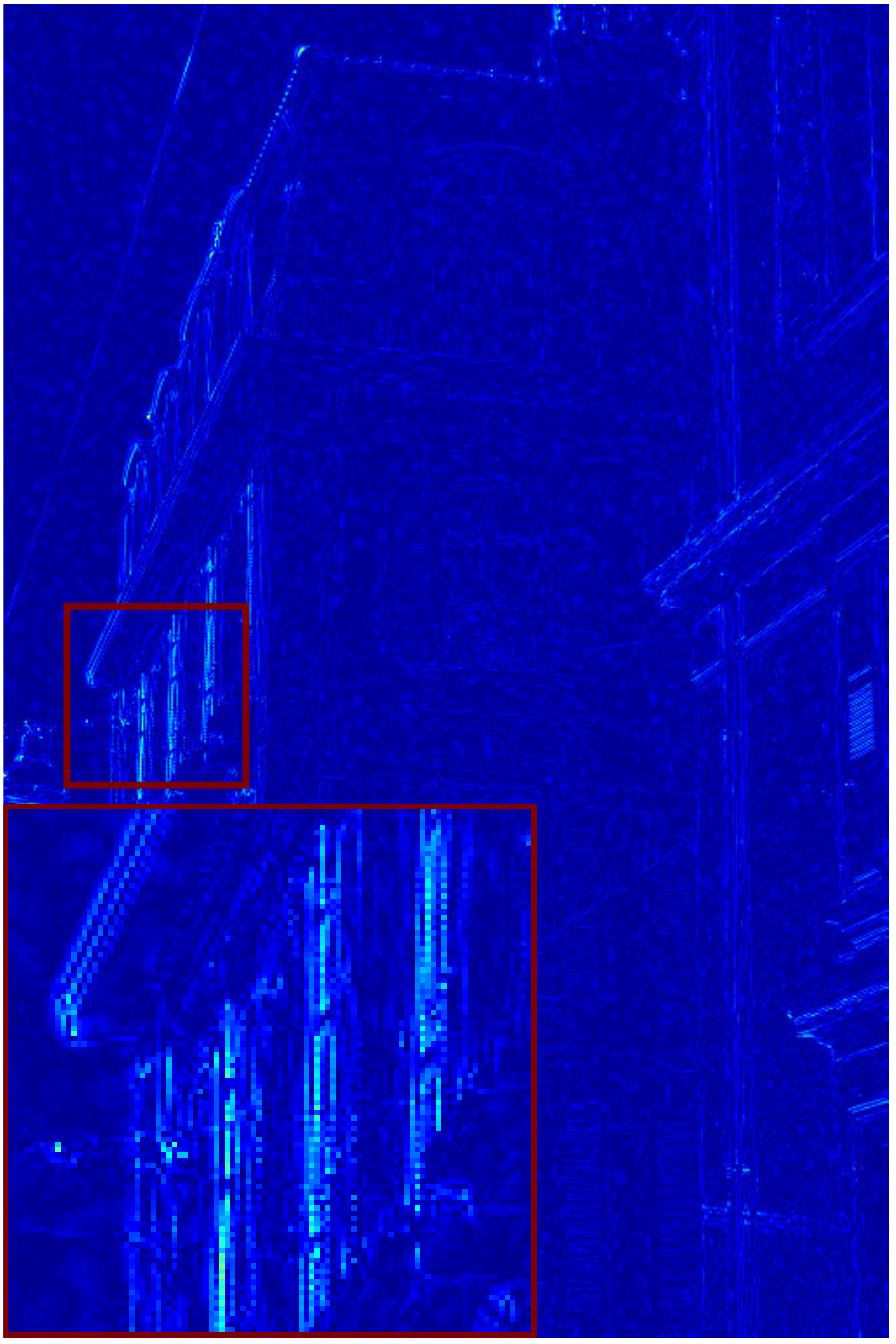}
		\end{minipage} 
		\begin{minipage}[b]{0.28\linewidth}
			\centering
			\includegraphics[width = 2.5cm, height = 2.5cm]{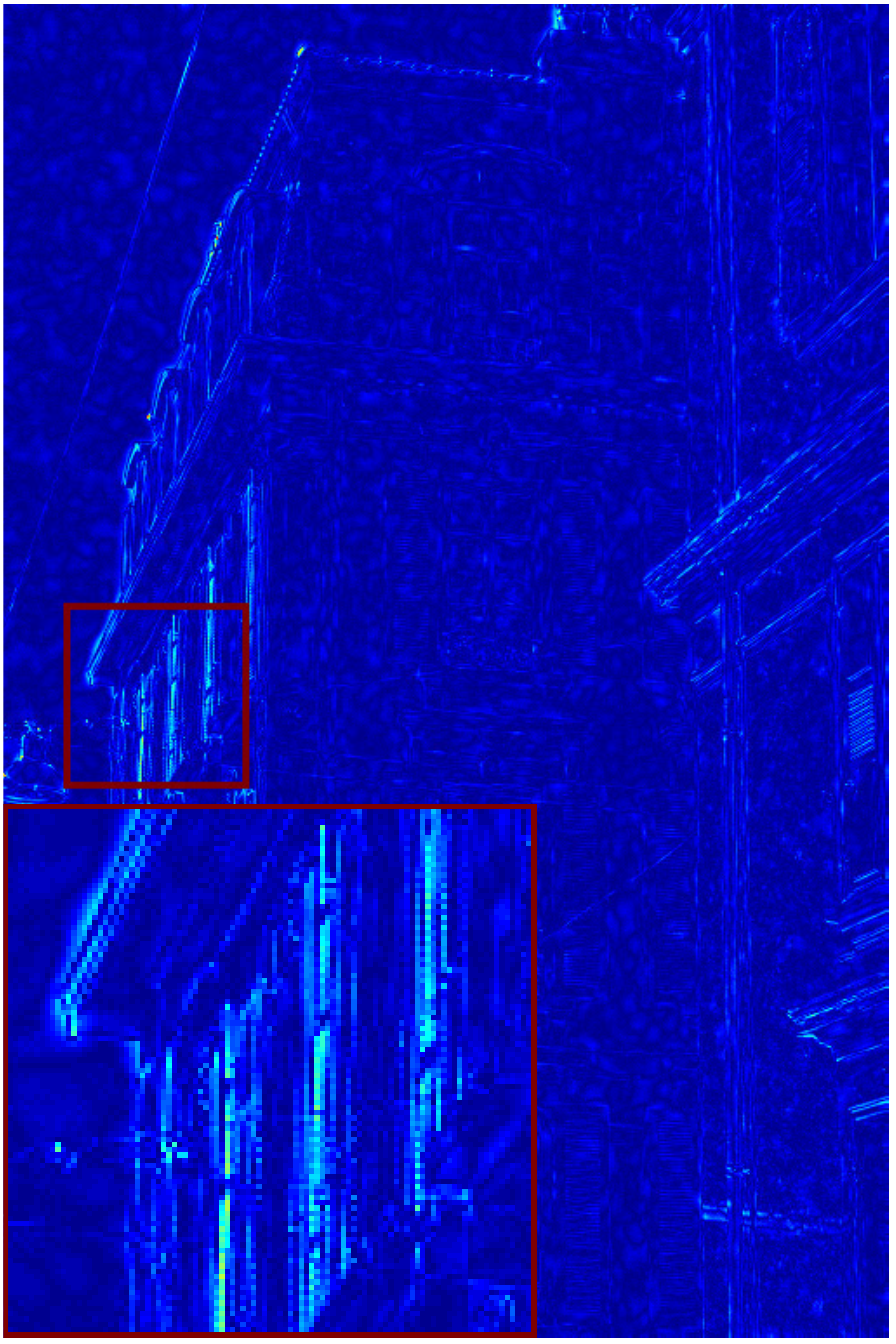}
		\end{minipage} 
		\begin{minipage}[b]{0.28\linewidth}
			\centering
			\includegraphics[width = 2.5cm, height = 2.5cm]{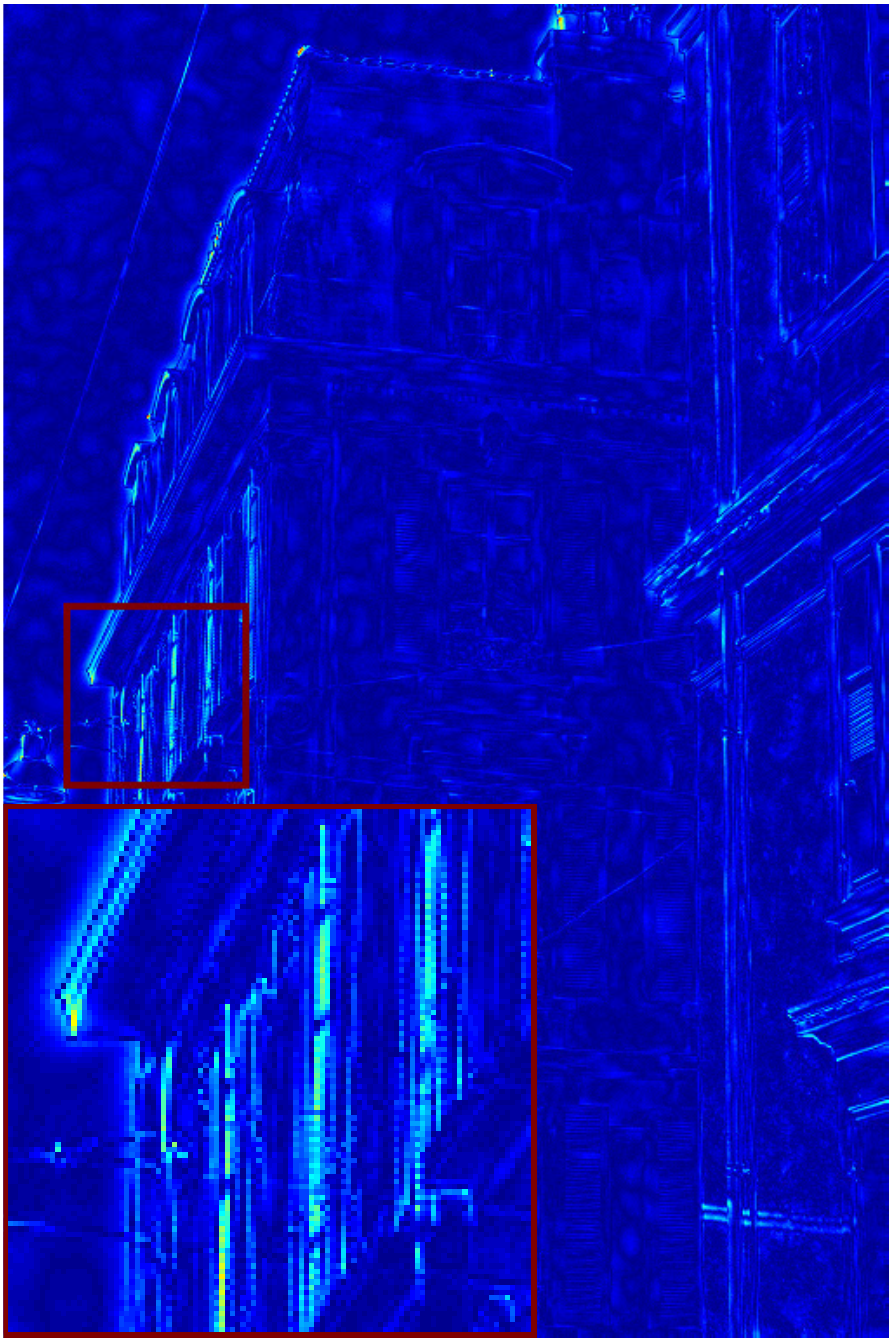}
		\end{minipage} 
		\\
		\begin{minipage}[b]{0.1\linewidth}
			\centering	
		\end{minipage}
		\begin{minipage}[b]{0.28\linewidth}
			\centering	{\footnotesize $\sigma=8$}
		\end{minipage} 
		\begin{minipage}[b]{0.28\linewidth}
			\centering	{\footnotesize $\sigma=16$} 
		\end{minipage} 
		\begin{minipage}[b]{0.28\linewidth}
			\centering	{\footnotesize $\sigma=24$} 
		\end{minipage}
		\\
		\begin{minipage}[b]{0.1\linewidth}
			\centering
			{\footnotesize DJF~\cite{li2016deep}} %
		\end{minipage} 
		\begin{minipage}[b]{0.28\linewidth}
			\centering
			\includegraphics[width = 2.5cm, height = 2.5cm]{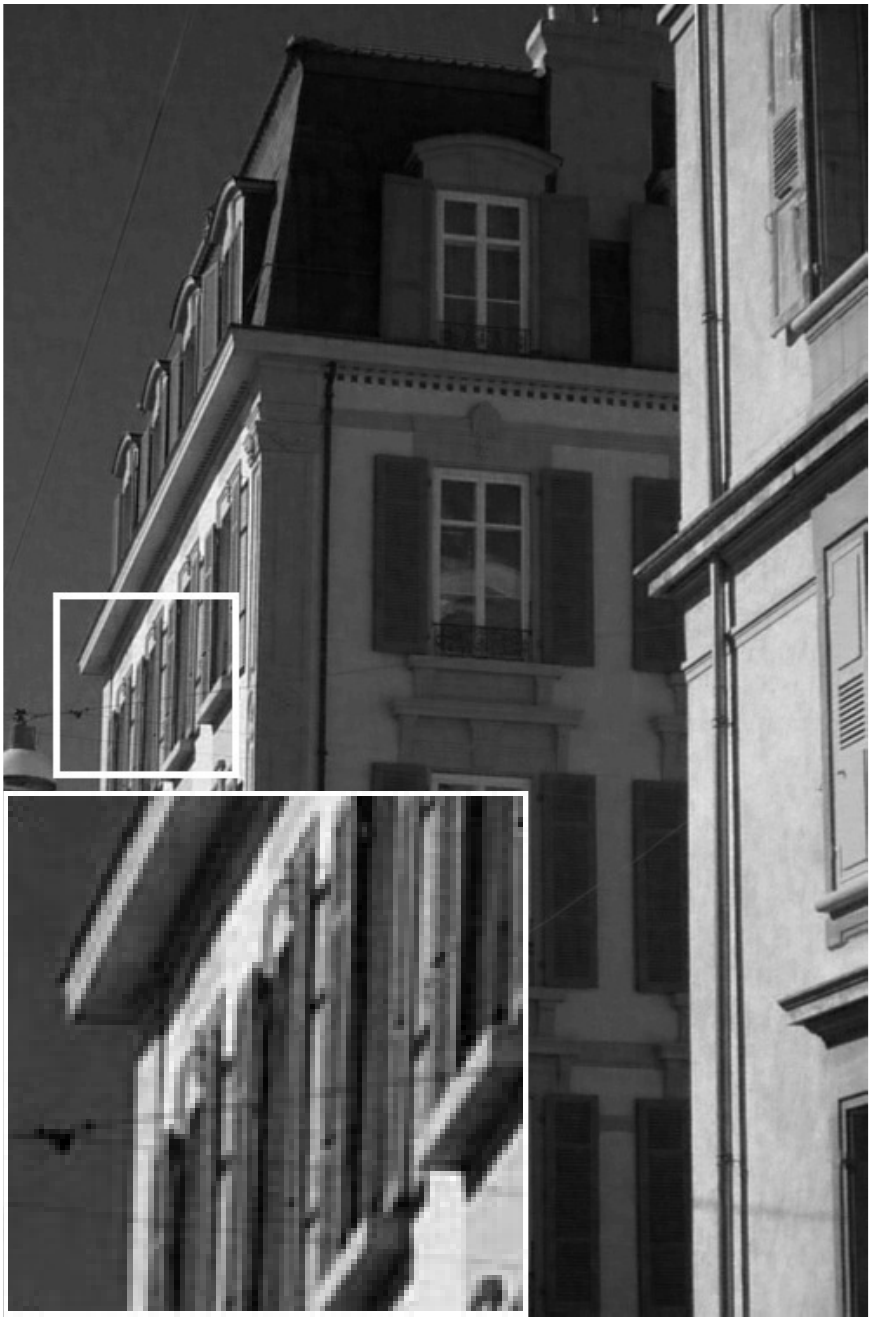}
		\end{minipage} 
		\begin{minipage}[b]{0.28\linewidth}
			\centering
			\includegraphics[width = 2.5cm, height = 2.5cm]{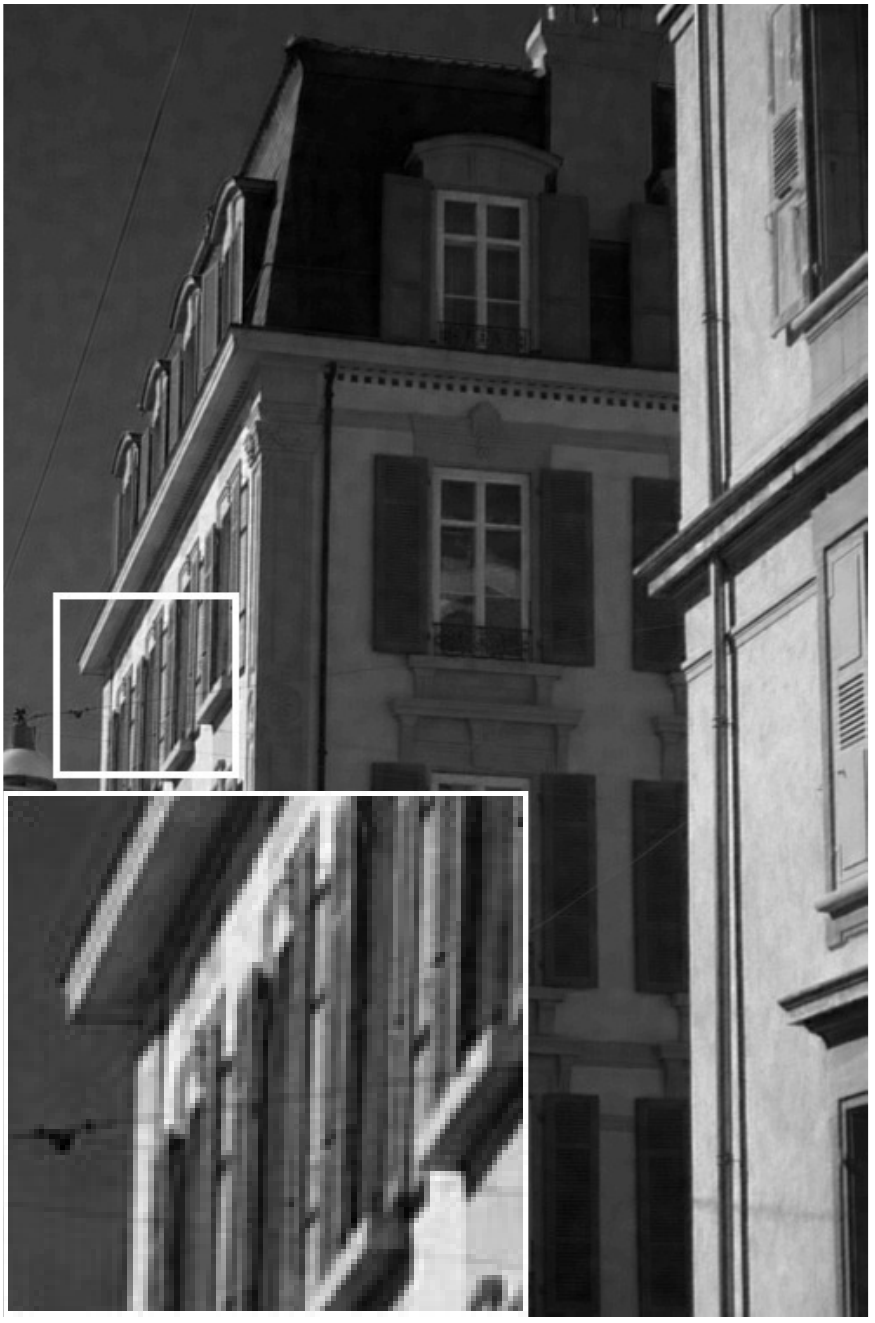}
		\end{minipage} 
		\begin{minipage}[b]{0.28\linewidth}
			\centering
			\includegraphics[width = 2.5cm, height = 2.5cm]{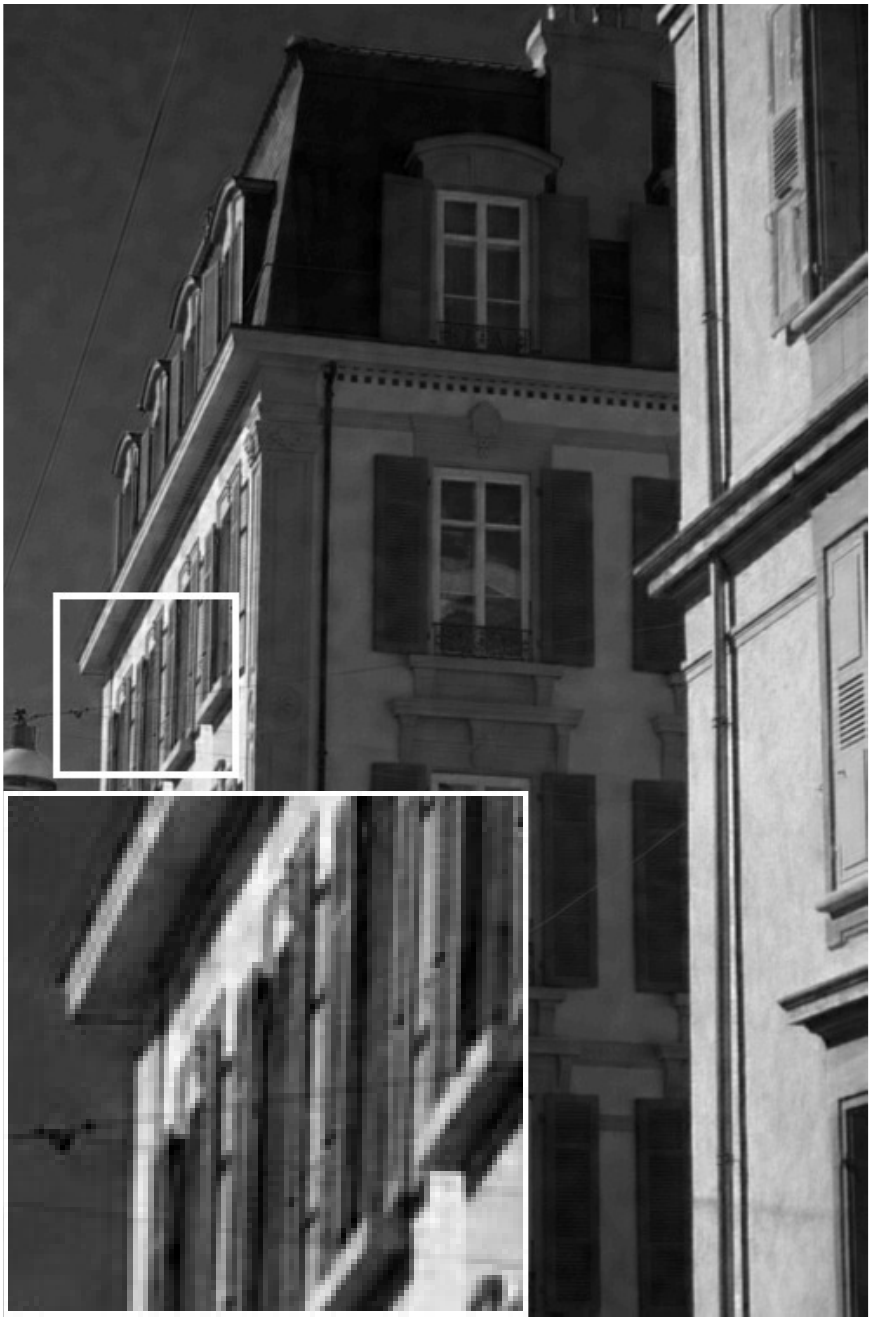}
		\end{minipage} 
		\\
		\begin{minipage}[b]{0.1\linewidth}
			\centering
			{\footnotesize ErrMap DJF~\cite{li2016deep}} %
		\end{minipage}  
		\begin{minipage}[b]{0.28\linewidth}
			\centering
			\includegraphics[width = 2.5cm, height = 2.5cm]{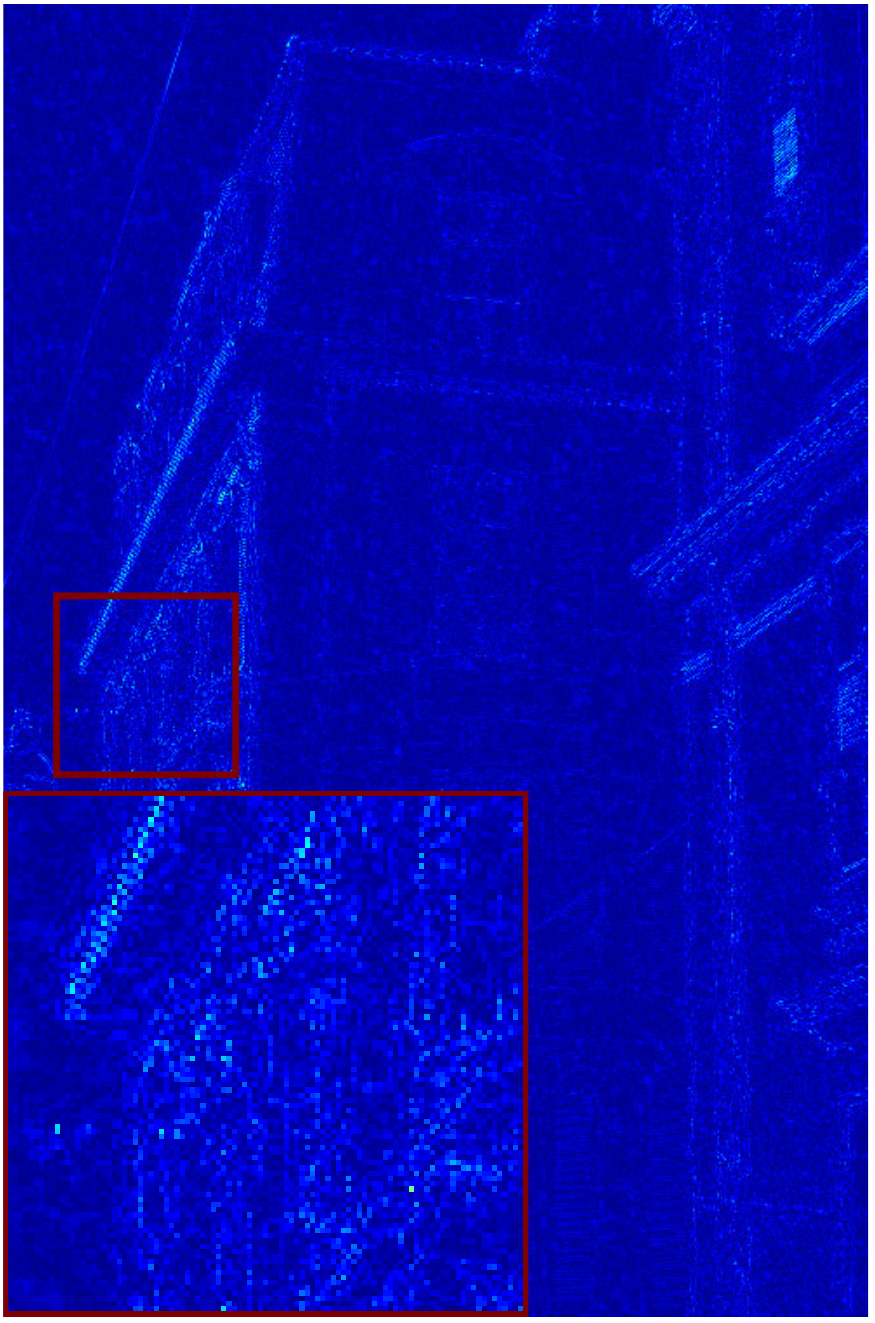}
		\end{minipage} 
		\begin{minipage}[b]{0.28\linewidth}
			\centering
			\includegraphics[width = 2.5cm, height = 2.5cm]{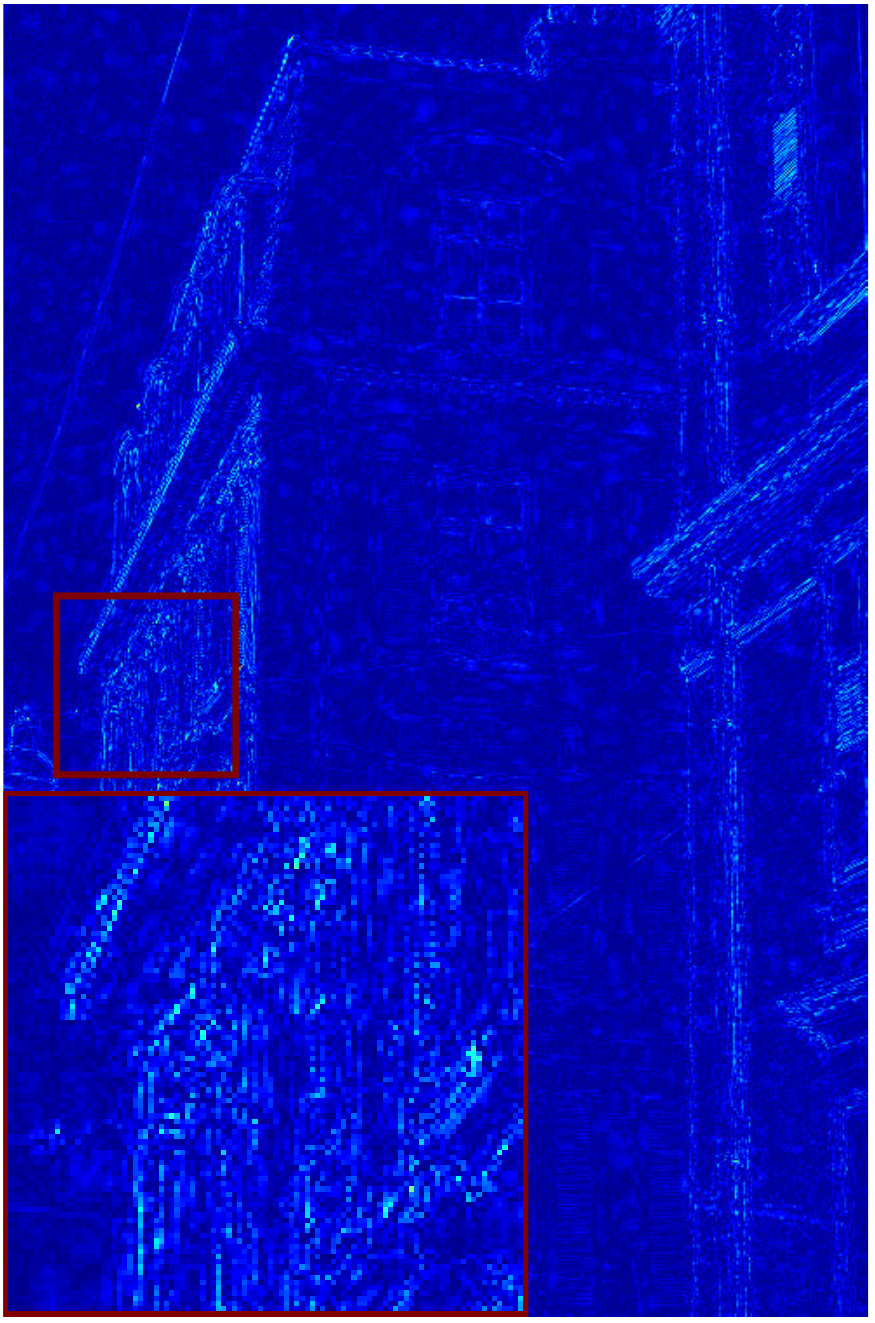}
		\end{minipage} 
		\begin{minipage}[b]{0.28\linewidth}
			\centering
			\includegraphics[width = 2.5cm, height = 2.5cm]{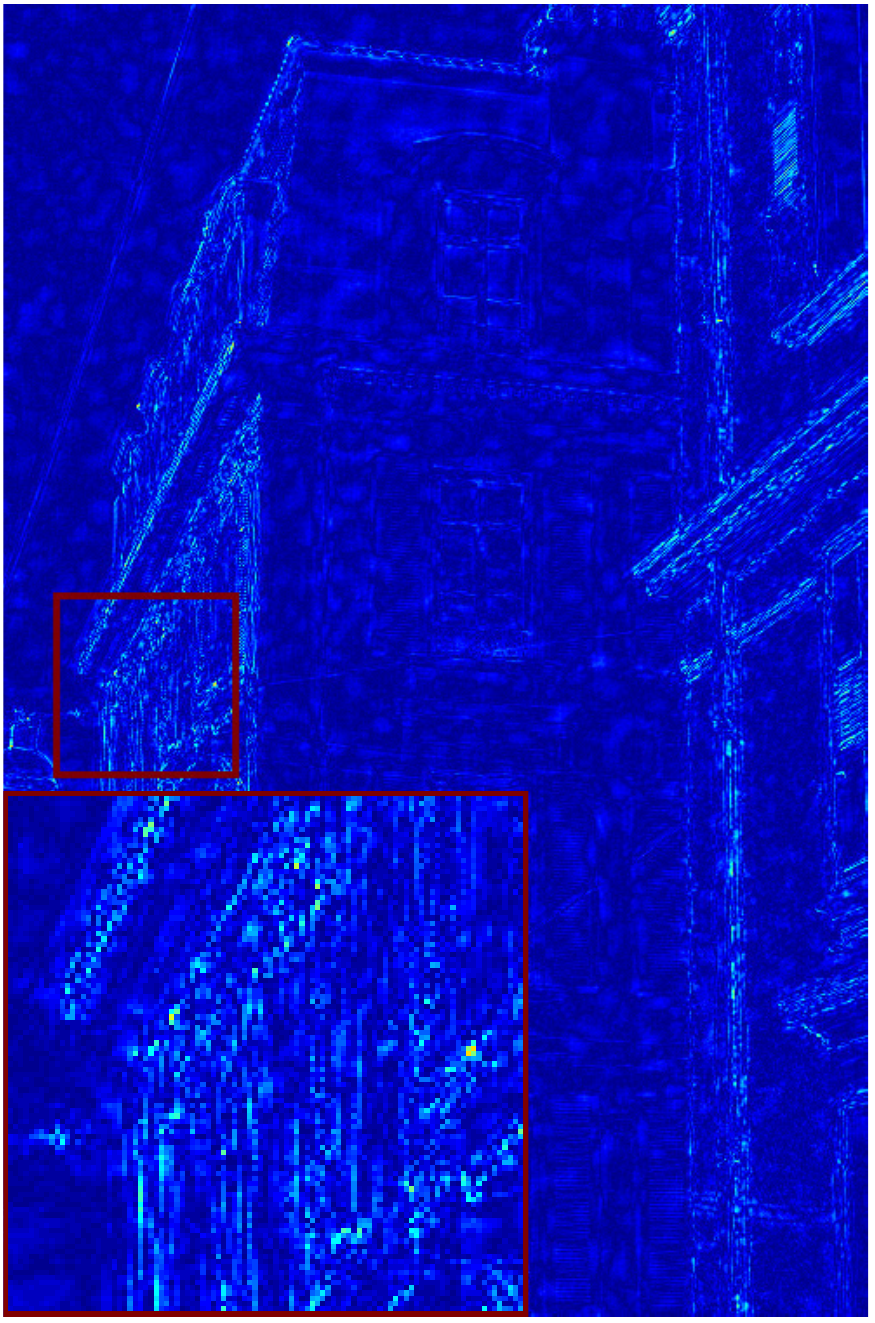}
		\end{minipage} 
		\\
		\begin{minipage}[b]{0.1\linewidth}
			\centering
			{\footnotesize Ours+}  
		\end{minipage} 
		\begin{minipage}[b]{0.28\linewidth}
			\centering
			\includegraphics[width = 2.5cm, height = 2.5cm]{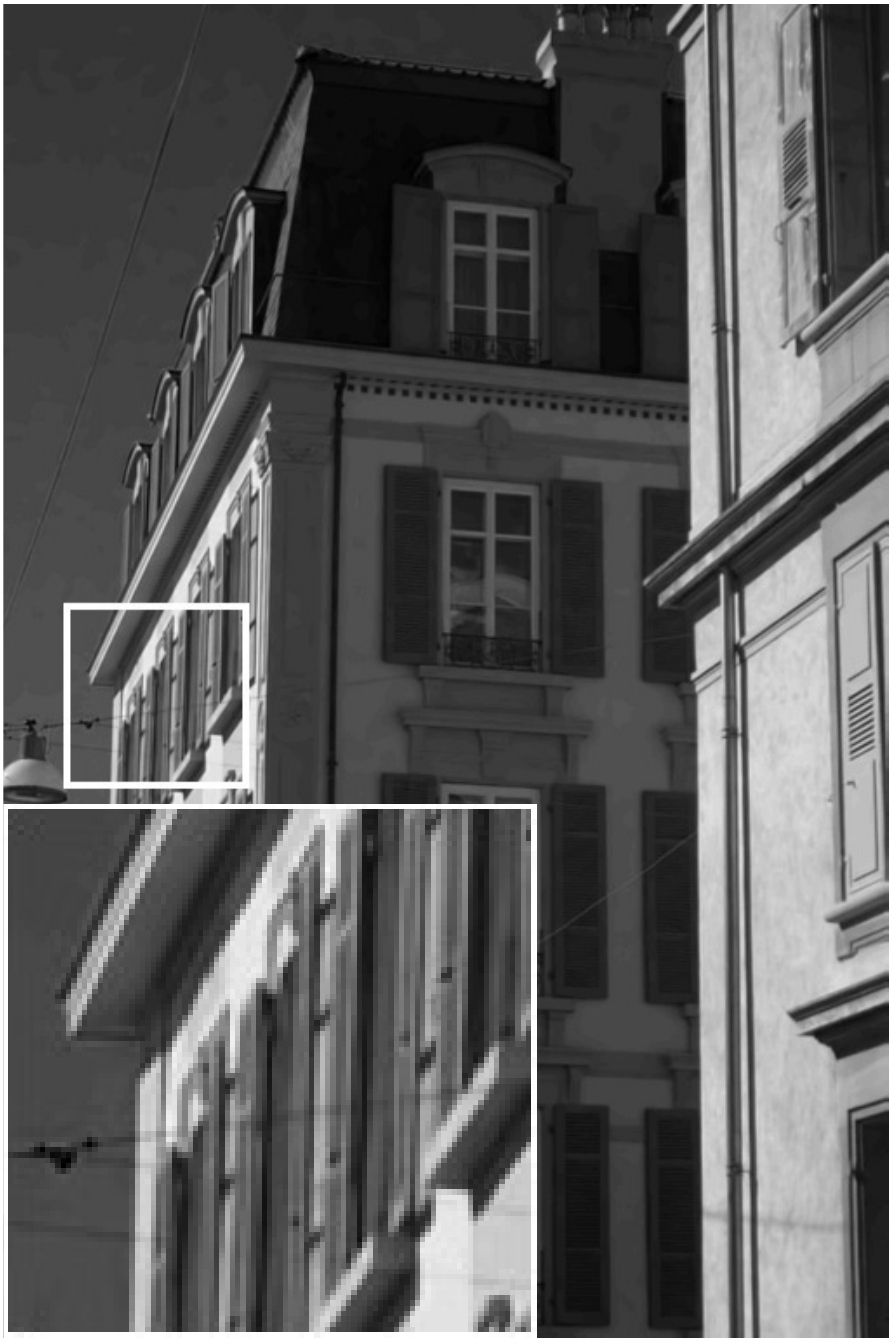}
		\end{minipage} 
		\begin{minipage}[b]{0.28\linewidth}
			\centering
			\includegraphics[width = 2.5cm, height = 2.5cm]{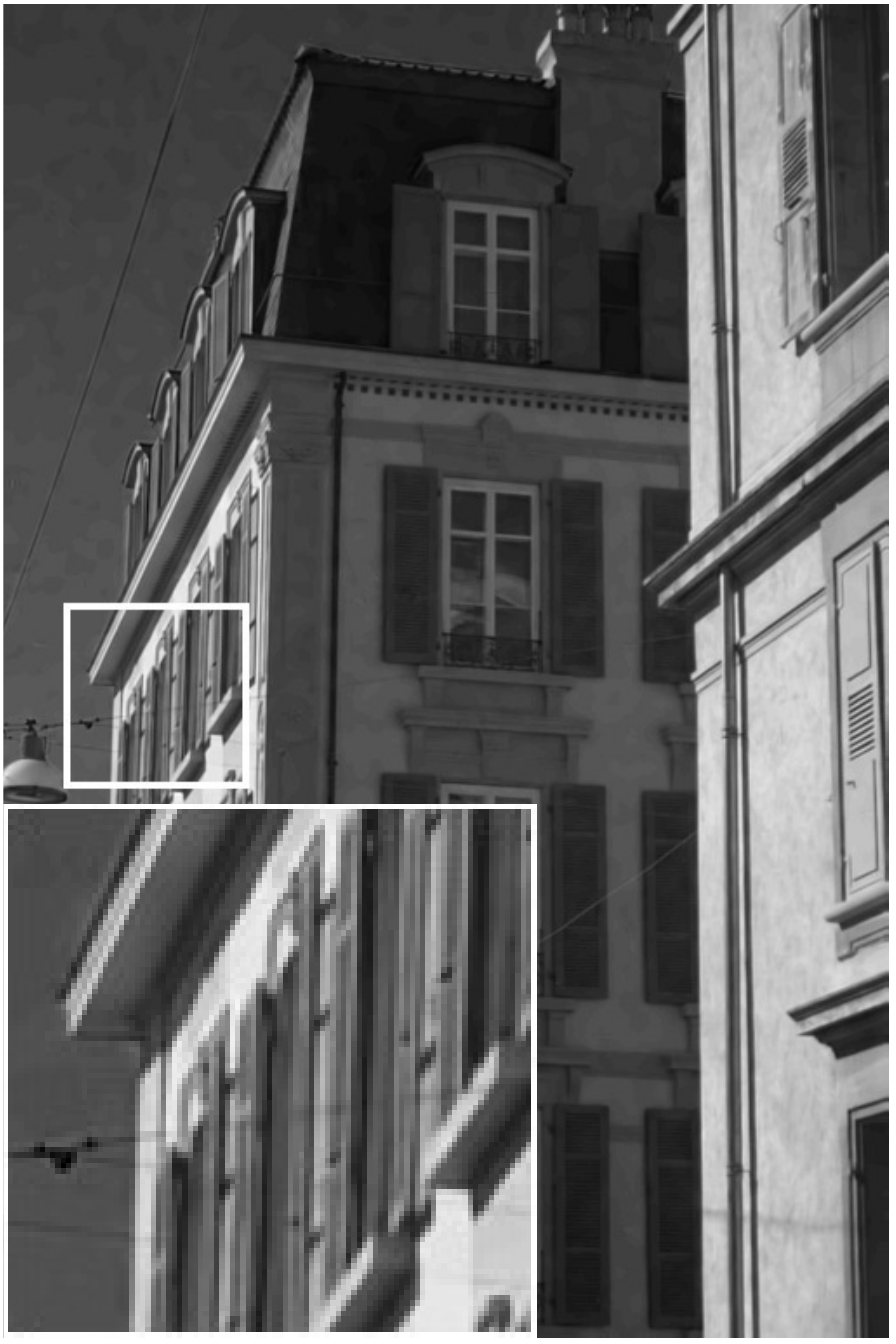}
		\end{minipage} 
		\begin{minipage}[b]{0.28\linewidth}
			\centering
			\includegraphics[width = 2.5cm, height = 2.5cm]{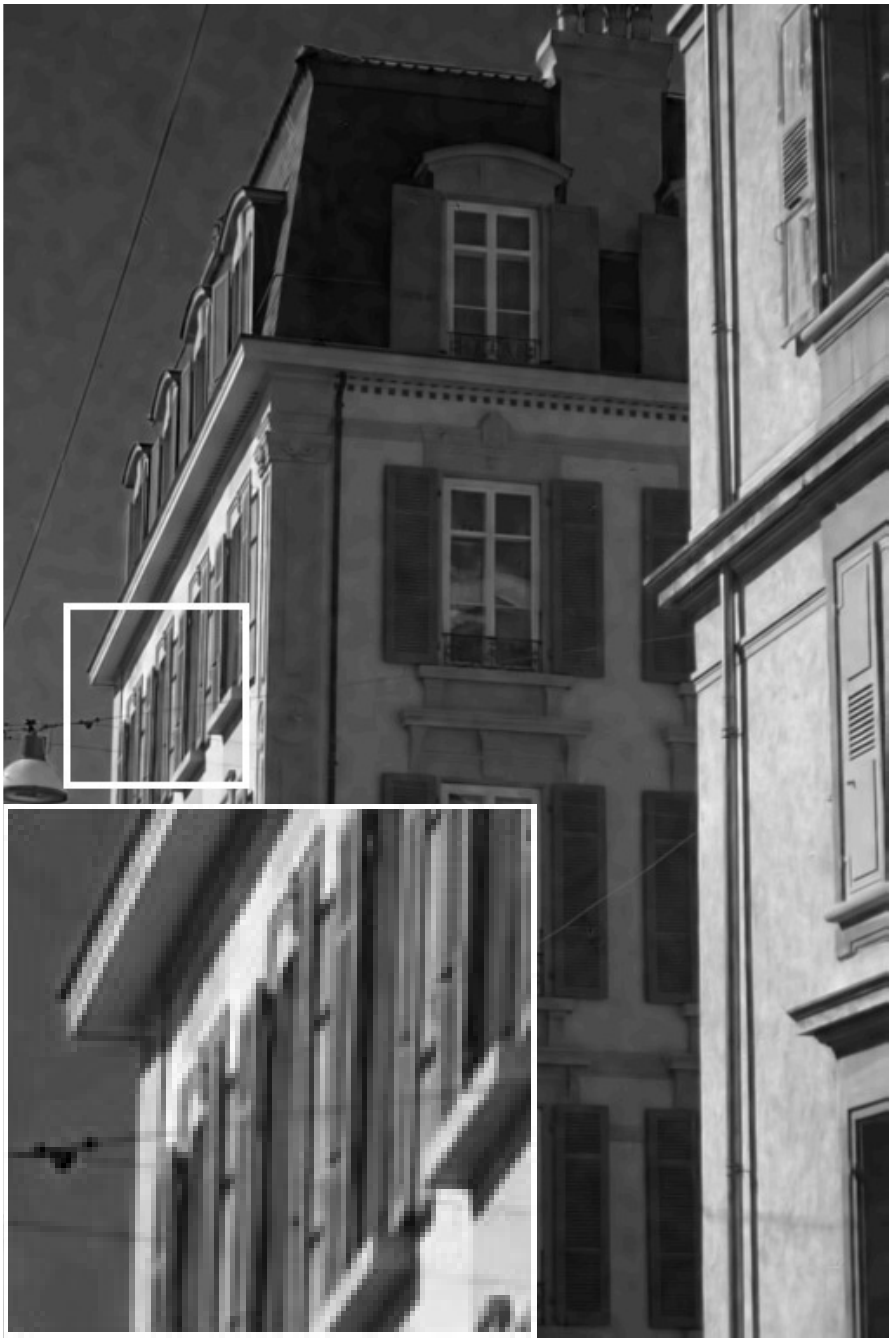}
		\end{minipage} 
		\\
		\begin{minipage}[b]{0.1\linewidth}
			\centering
			{\footnotesize ErrMap Ours+}  
		\end{minipage} 
		\begin{minipage}[b]{0.28\linewidth}
			\centering
			\includegraphics[width = 2.5cm, height = 2.5cm]{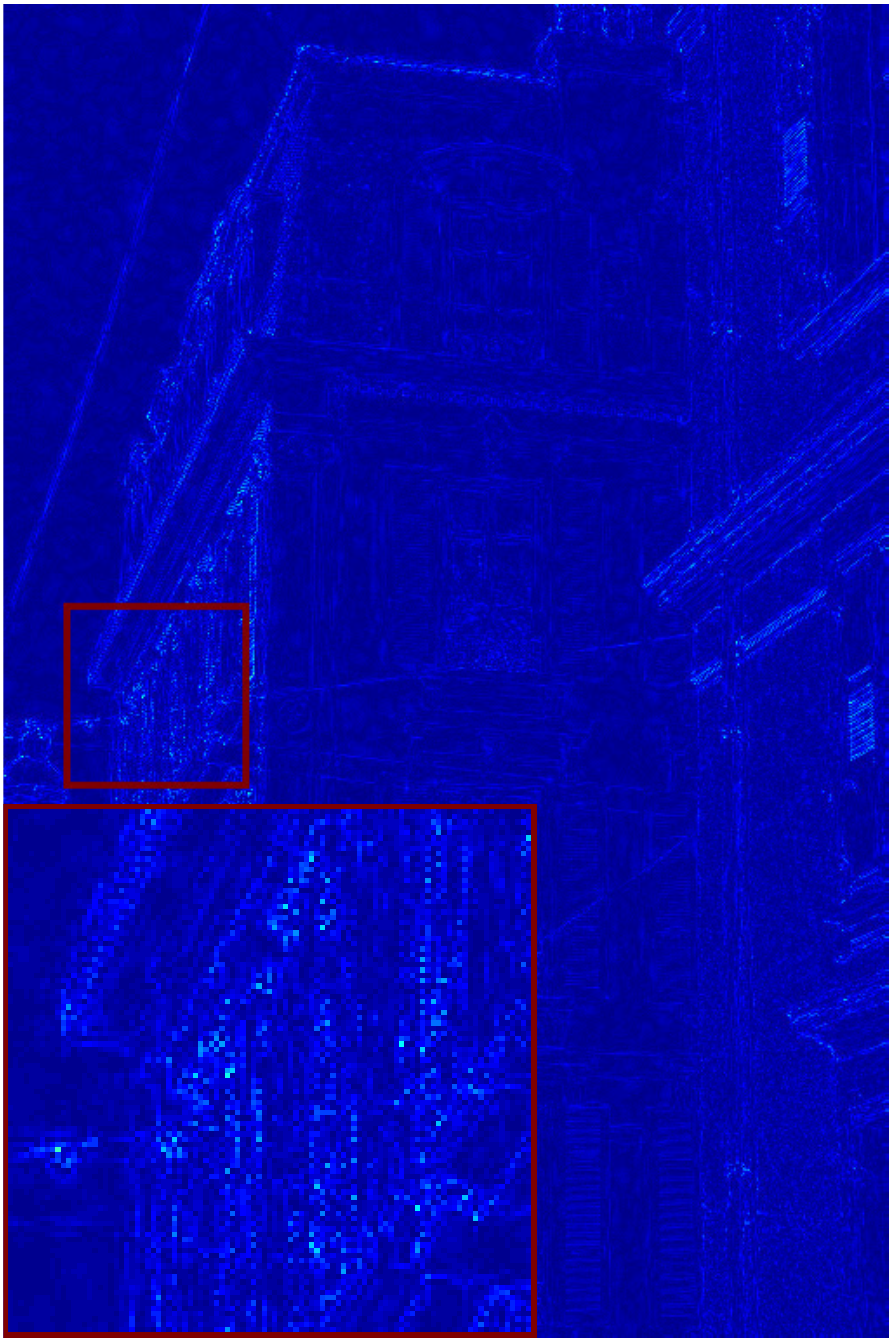}
		\end{minipage} 
		\begin{minipage}[b]{0.28\linewidth}
			\centering
			\includegraphics[width = 2.5cm, height = 2.5cm]{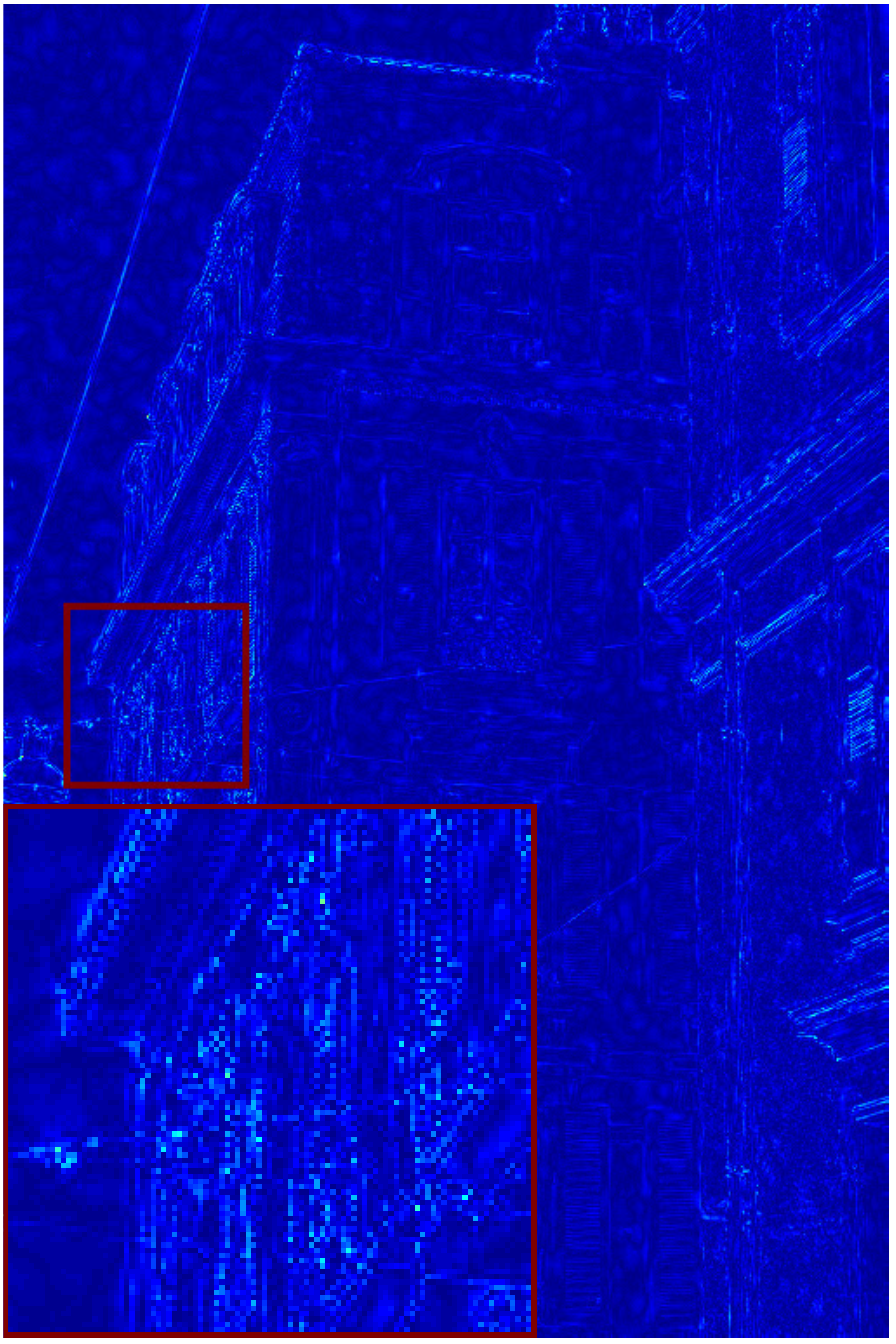}
		\end{minipage} 
		\begin{minipage}[b]{0.28\linewidth}
			\centering
			\includegraphics[width = 2.5cm, height = 2.5cm]{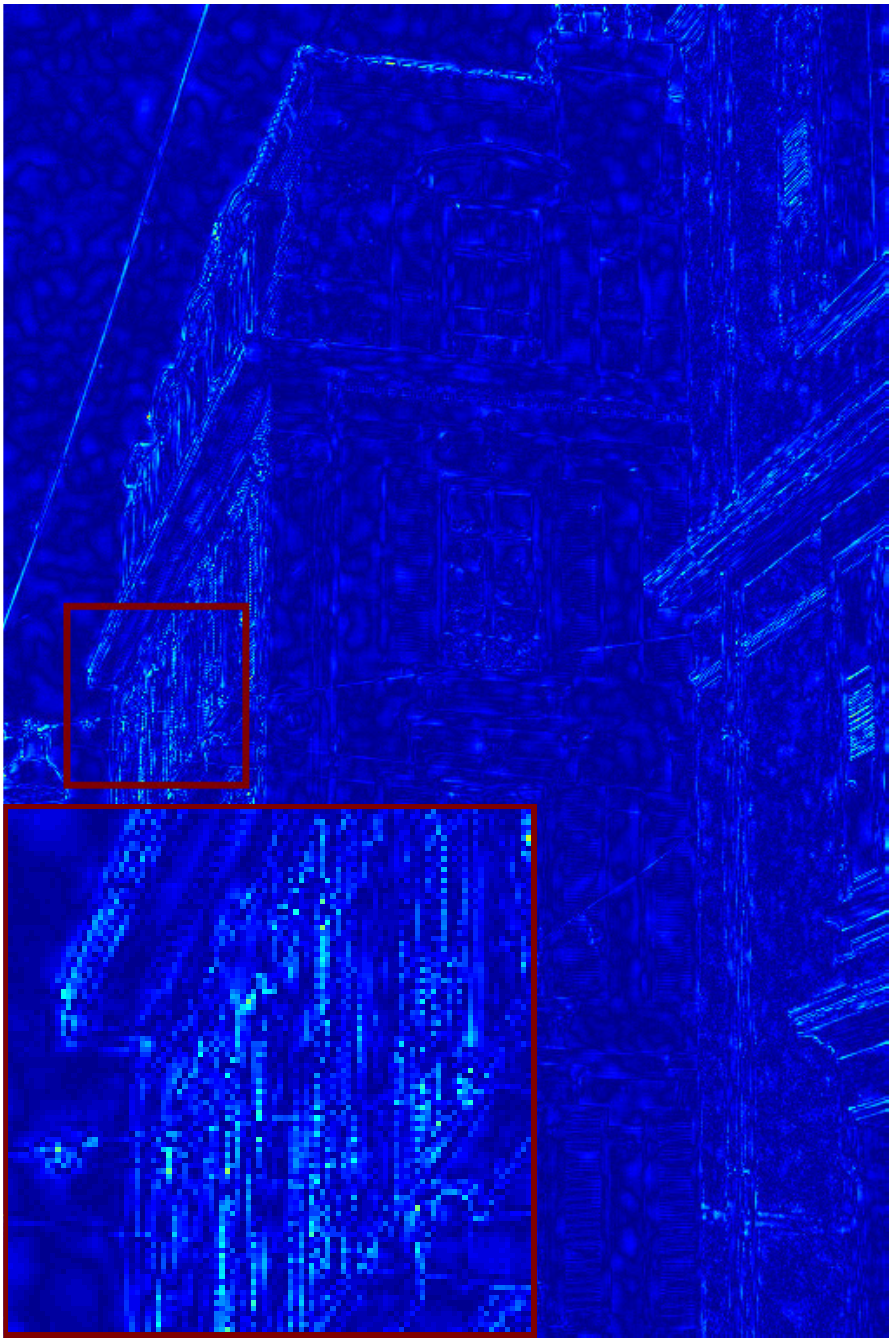}
		\end{minipage} 
		\\
		\begin{minipage}[b]{0.1\linewidth}
			\centering	
		\end{minipage}
		\begin{minipage}[b]{0.28\linewidth}
			\centering	{\footnotesize $\sigma=8$}
		\end{minipage} 
		\begin{minipage}[b]{0.28\linewidth}
			\centering	{\footnotesize $\sigma=16$} 
		\end{minipage} 
		\begin{minipage}[b]{0.28\linewidth}
			\centering	{\footnotesize $\sigma=24$} 
		\end{minipage}
	\end{multicols}	
	
	\vspace{-0.8cm}
	
	\begin{multicols}{1}  
	\begin{minipage}[b]{0.98\linewidth}
		{\footnotesize Colorbar}
		\includegraphics[width = 16.8cm, height=0.5cm]{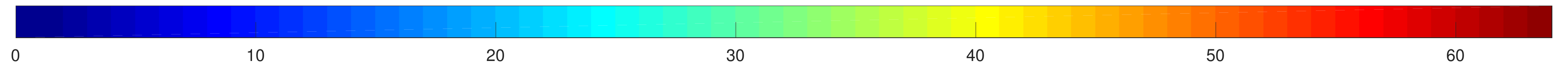}
	\end{minipage}
	\end{multicols}	

	\vspace{-0.6cm}

	\caption{Visual comparison for multimodal image denoising.}
	\label{Fig:DenoisedIms}
\end{figure*}


\section{Experiments}
\label{sec:Experiments}

\vspace{-0.2cm}

We now present a series of experiments to validate the effectiveness of the proposed multimodal image denoising approach. The dataset is from the EPFL infrared/RGB image database~\cite{brown2011multi}\footnote{\url{http://ivrl.epfl.ch/supplementary_material/cvpr11/}}. Each pair of infrared image and RGB image in the dataset have been registered with each other. The target modality is near infrared images and the guidance modality is RGB images. As the response of near infrared band exhibit poor correlation with the response of the visible band, it is usually difficult to infer the brightness of an infrared image given a corresponding RGB version. Thus, it is challenging to take good advantage of the RGB modality to purify the noisy infrared version.
%
%
%
The image pairs are randomly separated into two disjointed groups: training group (21 images) and testing group (8 images). We add zero-mean white Gaussian noise with different standard deviations $\sigma = [4,8,12,16,20,24]$ into the testing infrared images to generate the noisy version.

We compare our approach with state-of-the-art joint image filtering approaches, including Joint Bilateral Filtering (JBF)\cite{kopf2007joint}, Guided image Filtering (GF)\cite{he2013guided}, Static/Dynamic Filtering (SDF)\cite{ham2017robust}, Deep Joint image Filtering (DJF)\cite{li2016deep} and Joint Filtering via optimizing a Scale Map (JFSM)\cite{shen2015multispectral}. The same guidance images as in our approach are leveraged for these comparison approaches.
We adopt the Peak Signal to Noise Ratio (PSNR) and the RMSE as the image quality evaluation metrics which are commonly used in the image denoising literature. 

\vspace{-0.2cm}

\subsection{Coupled dictionary learning}
We perform coupled dictionary learning on a clean dataset, i.e., a corpus of clean multimodal image patch pairs extracted from the clean training images. We adopt a common operation to construct a patch-based training dataset. First, the clean infrared images and the corresponding RGB images with only intensity information are divided into a set of $\sqrt{n} \times \sqrt{n}$ patch pairs. Then, we remove the mean from each patch, as the DC component is always preserved well during the denoising process. Finally, we vectorize these patches to form the training datasets $\{(\mathbf{x}_i, \mathbf{y}_i) \}_{i=1}^T$ of dimension $n \times T$. Once the training dataset is prepared, we apply our coupled dictionary learning algorithm to learn the dictionary pairs $[\boldsymbol{\Psi}_{c},\boldsymbol{\Psi}]$ and $[\boldsymbol{\Phi}_{c},\boldsymbol{\Phi}]$ from the training datasets. The parameter setting is as follows: patch size $ n = 8 $, dictionary size $k= 1024$, training size $T \approx 50,000$, $\lambda = 0.05$. 


Figure~\ref{Fig:LearnedD} shows the learned coupled dictionaries from the corpus of clean infrared images and corresponding RGB version. We can find that any pair of atoms from common dictionaries $\boldsymbol{\Psi}_{c}$ and $\boldsymbol{\Psi}_{c}$ capture associated edges, blobs, textures with the same direction and location, as well as exhibit considerable resemblance and strong correlation to each other. This outcome indicates that the common dictionaries have indeed captured the similarities between infrared and RGB modalities. In contrast, the learned unique dictionaries $\boldsymbol{\Psi}$ and $\boldsymbol{\Phi}$ represent the disparities of these modalities and therefore rarely exhibit resemblance.

\vspace{-0.2cm}

\subsection{Coupled image denoising}

During the coupled image denoising phase, we evaluate basic approach with standard sparsity regularization and advanced version using group sparsity regularization to incorporate self-similarity prior.
%
%
%
Figure~\ref{Fig:PSNR_RMSE} and \ref{Fig:DenoisedIms} demonstrate the denoising performance, visual quality of the purified infrared images, as well as the corresponding error maps. As shown in these figures, our approach substantially attenuates the noise and, at the same time, reliably reserves image sharp details and suppresses artifacts. Therefore, the purified infrared images by our approach are cleaner and more visually plausible than the reconstruction by the state-of-the-art joint image filtering approaches. The visual quality is also demonstrated by the error maps where the denoised infrared images using our approach exhibit the least residual for different noise levels in comparison with the competing methods. In particular, it indicates that detailed structure information such as sharp edges, textures and stripes, can be effectively captured by learned coupled dictionaries. The average PSNR and average RMSE results for the multimodal image denoising task, shown in Table~\ref{Tab:PSNR_RMSE} and Figure~\ref{Fig:PSNR_RMSE}, also confirm that our approach exhibits notable advantages over the competing methods.\footnote{
	More results can be found in the appendix of supplementary materials.}


\vspace{-0.2cm}

\section{Conclusion}
\label{sec:Conclusion}

\vspace{-0.2cm}

This paper proposes a new effective multimodal image denoising approach based on coupled dictionary learning. The proposed approach explicitly incorporates sparsity prior, self-similarity prior and cross-similarity prior in the data model to captures the similarities and disparities between different image modalities in a learned sparse feature domain in \emph{lieu} of the original image domain. This scheme is able to exploit a guidance image to aid the denoising of the target image of interested modality, achieving notable benefits in the task with respect to the state-of-the-art.

%
%




\clearpage
\pagebreak

\end{document}